\pdfoutput=1

\documentclass[11pt]{article}

\usepackage[]{latex/acl}

\usepackage{times}
\usepackage{latexsym}

\usepackage[T1]{fontenc}

\usepackage[utf8]{inputenc}

\usepackage{microtype}

\usepackage{inconsolata}

\usepackage{colortbl}  
\usepackage[table]{xcolor}
\usepackage{subfigure}
\usepackage{multirow}
\usepackage{newfloat}
\usepackage{listings}
\usepackage{makecell}
\usepackage{booktabs} 
\usepackage{amsfonts} 
\usepackage{amsmath}  
\usepackage{mdframed} 
\usepackage{comment}
\usepackage{bbding}
\usepackage{environ}
\usepackage{tikz}
\usepackage{enumitem}
\usepackage{indentfirst}
\usepackage{floatrow}
\usepackage{bbm}
\usepackage{graphicx}
\usepackage{subfig}
\usepackage{algorithm}
\usepackage{algorithmic}
\usepackage{xcolor}
\usepackage{tcolorbox}
\usepackage{xcolor}
\usepackage{colortbl}

\title{Weed Out, Then Harvest: Dual Low-Rank Adaptation is an Effective Noisy Label Detector for Noise-Robust Learning}

\author{
Bo Yuan, Yulin Chen, Yin Zhang\thanks{Corresponding Author}\\
  Zhejiang University, Hangzhou, China \\
  \texttt{\{byuan,yulinchen,yinzh\}@zju.edu.cn} \\
}
 
\begin{document}
\maketitle
\begin{abstract}


Parameter-efficient fine-tuning (PEFT) large language models (LLMs) have shown impressive performance in various downstream tasks. However, in many real-world scenarios, the collected training data inevitably contains noisy labels. To learn from noisy labels, most solutions select samples with small losses for model training. However, the selected samples, in turn, impact the loss computation in the next iteration. An inaccurate initial selection can create a vicious cycle, leading to suboptimal performance. To break this cycle, 
 we propose Delora, a novel framework that decouples the sample selection from model training. 
 For sample selection, Delora establishes a noisy label detector by introducing clean and noisy LoRA. Benefiting from the memory effect, the clean LoRA is encouraged to memorize clean data, while the noisy LoRA is constrained to memorize mislabeled data, which serves as a learnable threshold for selecting clean and noisy samples. 
For model training, Delora can use carefully selected samples to fine-tune language models seamlessly. Experimental results on synthetic and real-world noisy datasets demonstrate the effectiveness of Delora in noisy label detection and text classification.

\end{abstract}

\section{Introduction}
LLMs are extremely powerful, yet they are expensive to train. By balancing performance with practicality, PEFT has become a popular technique for adapting the LLM for downstream applications. Notable PEFT methods include Low-Rank Adaptation (LoRA) \cite{DBLP:conf/iclr/HuSWALWWC22}, adaptors \cite{DBLP:conf/icml/HoulsbyGJMLGAG19}, and prompts \cite{DBLP:conf/acl/LiuJFTDY022}. Instead of fine-tuning all weights, PEFT fixes the backbone model parameters while adding a few learnable parameters for adaptation. While promising, PEFT techniques rely on perfectly labeled datasets, which may not be readily available in many real-world applications, limiting their broader application.


\begin{figure}[t!]
\centering
\includegraphics[width=7.7cm,height=1.5cm]{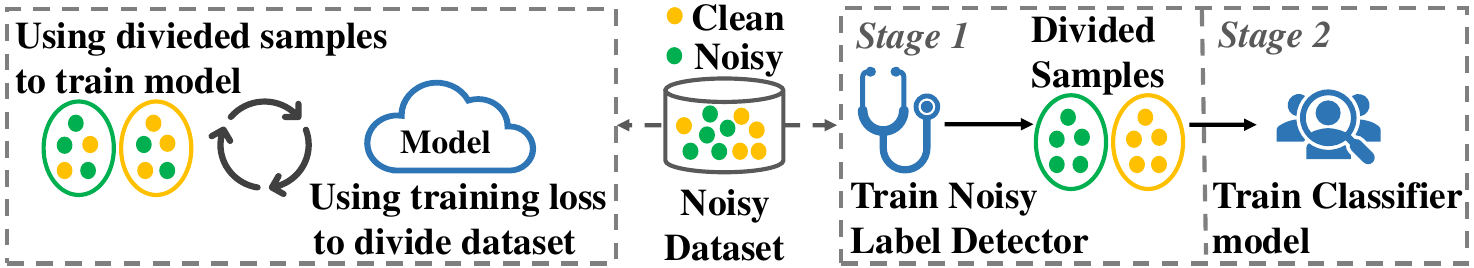}
\caption{A comparison between other sample selection methods (left) and our method (right) for LNL tasks. Our method decouples sample selection  (stage 1) from model training (stage 2) by training a noisy label detector and classifier model separately.}
\label{compare_others_methods}
\end{figure}

To tackle this issue, a recent work \cite{DBLP:conf/acl/KimK024} explores PEFT methods on imperfectly labeled datasets ($\textit{i.e.}$, datasets with noisy labels) to learn with noisy labels (LNL). It uses training losses to select clean data and suggests a routing-based PEFT method that preferentially trains PEFT modules on clean data. Similar to previous LNL methods \cite{DBLP:conf/nips/HanYYNXHTS18, DBLP:conf/nips/ShuXY0ZXM19,DBLP:conf/coling/QiaoDDLCC022, DBLP:conf/acl/YuanCZJ24} based on sample selection, they both adopt the "small-loss" mechanism to select clean data because the model tends to fit clean samples earlier than noisy samples during training. However, these methods are inherently affected by the label noise, as losses used for sample selections are extracted from the model that is being trained. Specifically, the sample selection process affects subsequent training, and training loss in turn influences the sample selection. If the initial sample selection is poor, it leads to an inescapable vicious cycle. We contend that in such a strategic feedback loop, a good sample selection can itself fall into a new paradox, akin to: \textit{\textbf{It's a catch-22 situation: A good sample selection requires generalizability, while generalizability requires sample selection.}}

Given this, let us leave the existing framework and revisit the LNL task. The main difficulty here is that we need to select clean samples to train a strong model while avoiding the effects of noisy samples. \textit{Can we, perhaps, decouple sample selection from the model training, making them independent of each other?} With this question in mind, we propose a new framework that firstly leverages PEFT modules to construct a noisy label detector for sample selection and then trains the model using selected samples, as shown in Figure \ref{compare_others_methods}.

In the first stage, we introduce two distinct PEFT modules ($\textit{i.e.}$, clean LoRA and noisy LoRA) to construct a noisy label detector. The parameters in clean LoRA are termed the ideal parameters used to memorize clean data, while parameters in noisy LoRA are called noise parameters used to memorize mislabeled data. Our noisy label detector can be derived from the cross-entropy between the predictions of clean/noisy LoRA on the training text and its corresponding labels. If the text exhibits a higher cross-entropy value with the noisy LoRA than the clean LoRA, it is considered a correctly labeled sample, $\textit{i.e.}$, a clean sample. This design draws inspiration from a recent study \cite{DBLP:journals/corr/abs-2412-04465} that decomposes an image into its constituent subject and style, represented as two distinct LoRAs. This insight has inspired us to explore different LoRAs to memorize clean and noisy samples separately in LNL. However, as we cannot directly obtain clean and noisy samples from datasets, controlling LoRAs to accomplish our notion is a challenge. Note that the memorizing effect \cite{DBLP:conf/icml/ArpitJBKBKMFCBL17} demonstrates that deep networks would first memorize clean samples and then noisy samples. Based on this, we introduce dynamic regularization to adjust the parameter updates of the two LoRAs over time. Specifically, we constrain noise parameter updates of noisy LoRA in the early training stage and make ideal parameters of clean LoRA completely memorize clean data. As training progresses, the restrictions on noisy LoRA are gradually lessened while the constraints on clean LoRA are reinforced, resulting in noisy samples being mostly memorized by noisy LoRA.


In the second stage, we train the classifier model using samples carefully selected through the noisy label detector. In this step, we first utilize selected clean samples as contextual references for reliable relabeling of noisy samples, and then merge the clean samples and relabeled noisy samples to fine-tune the classifier model. In this way, we can obtain a denoised fine-tuning dataset while maximizing data utility to improve the generalization ability of our framework in LNL tasks. It is worth noting that the sample selection is independent of the in-training classifier model, which can effectively avoid the issue of vicious cycles. Overall, our main contributions can be summarized as follows:

\begin{itemize}
\setlength{\itemsep}{1pt}
\setlength{\parsep}{1pt}
\setlength{\parskip}{1pt}
\item We propose a new framework that decouples sample selection from model training
to address the LNL tasks, effectively avoiding the vicious cycles common in existing solutions.
\item For sample selection, we introduce two LoRAs to construct a noisy label detector: the ideal parameters of clean LoRA to memorize clean samples and the noise parameters of noisy LoRA to memorize noisy samples.
\item Based on the memory effect, we dynamically constrain the parameters of LoRAs so that noisy LoRA absorbs the side effects of noisy samples and clean LoRA fits clean samples.

\item We conduct extensive experiments across diverse text classification datasets under varying noise conditions, demonstrating the superiority of  Delora over existing baselines in both noise label detection and text classification.

\end{itemize}

\section{Related Work}
\subsection{Sample Selection for LNL.}
Model fine-tuning often relies on large-scale, high-quality data. However, the dataset inevitably introduces noisy labels during collection. For LNL, sample selection methods are popular solutions that strive to select clean samples from noisy datasets using specific criteria, such as the widely applied "small-loss" mechanism or model predictions. Among them, \cite{DBLP:conf/nips/HanYYNXHTS18, DBLP:conf/nips/ShuXY0ZXM19,DBLP:conf/coling/QiaoDDLCC022} set a fixed threshold for loss value to divide the noisy data, \citet{DBLP:conf/acl/YuanCZJ24} further proposes a dynamic-enhanced threshold strategy to improve the previous method based on fixed thresholds. However, they all require manually setting thresholds, which increases the cost of hyper-parameter tuning. For our proposed Delora, the noisy LoRA prediction functions as a learnable "threshold" for identifying noisy labels. Moreover, these approaches remain unstable and susceptible to vicious cycles (self-confirmation bias), particularly in high-noise scenarios. This instability stems from their inherent reliance on the in-training model \cite{DBLP:conf/mm/FengTP24} and their exclusive focus on learning with label noise from scratch. Different from previous studies, our Delora decouples sample selection and classifier model training to break this vicious cycle.

\subsection{Parameter-Efficient Fine-tuning.}
As LLMs get bigger, PEFT is more essential for conserving resources. A lot of strategies, including LoRA, adaptors, and prompt learning, have been put forth by researchers to improve fine-tuning effectiveness. Among them, the PEFT based on LoRA \cite{DBLP:conf/iclr/HuSWALWWC22} is popular and widely used. Recently, \citet{DBLP:conf/acl/KimK024} explores PEFT's robustness to noisy labels and finds that PEFT's limited capacity enhances robustness to noisy samples but also hinders learning from clean samples. Then, they propose a routing-based method to adaptively activate PEFT modules. However, they still require manually setting a fixed loss threshold and relying on losses of the in-training model for sample selection. We address their shortcomings and leverage PEFT's limited capacity to separately memorize clean and noisy samples.

\section{Preliminaries}
\label{Preliminaries}
\textbf{Problem Setup.} Given a training dataset $\mathcal{D}$=$\left\{(x_i,y_i)\right\}^{N}_{i=1}$ with $N$ samples and $K$ classes, where $y \in \left\{1,\dots,K\right\}$ is the observed label of the sample $x$, and $y$ is possibly the incorrect label.

\textbf{LoRA fine-tuning.} LoRA (Low-Rank Adaptation) is a PEFT (parameter-efficient fine-tuning) technique for effectively modifying LLMs for a new downstream task. The main idea behind LoRA is that while fine-tuning, the weight updates $\Delta w$  to the base model weights $w_0 \in \mathbbm{R}^{m \times n}$ have a low intrinsic rank. As a result, the update $\Delta w$ may be broken down into two low-rank matrices, $B 
\in \mathbbm{R}^{m \times r}$
and $A \in \mathbbm{R}^{r \times n}$, for efficient parameterization, with $\Delta w=BA$. For $r$, it means the intrinsic rank of 
$\Delta w$ with $r \ll min(m,n)$. During training, only $A$ and $B$ are updated to find suitable $\Delta w=BA$ targeting specific tasks while keeping $w_0$ constant. For inference, the updated weight matrix $w$ can be obtained as $w = w_0 + \Delta w$. Denoting the prediction with LoRA modules as $f(x, w_0 + \Delta w)$, the objective with an arbitrary loss function $\mathcal{L}$ can be formulated as follows:
\begin{align}
\min_{\Delta w} \mathcal{L}(x) = \mathcal{L}(f(x, w_0  + \Delta w), y),
\end{align}
where $x$ and $y$ are the training sample and its label, respectively. The model $f$ with LoRA modules $\Delta w$ is only updated during training.

\begin{figure*}[t]
\centering
\includegraphics[width=15.5cm]{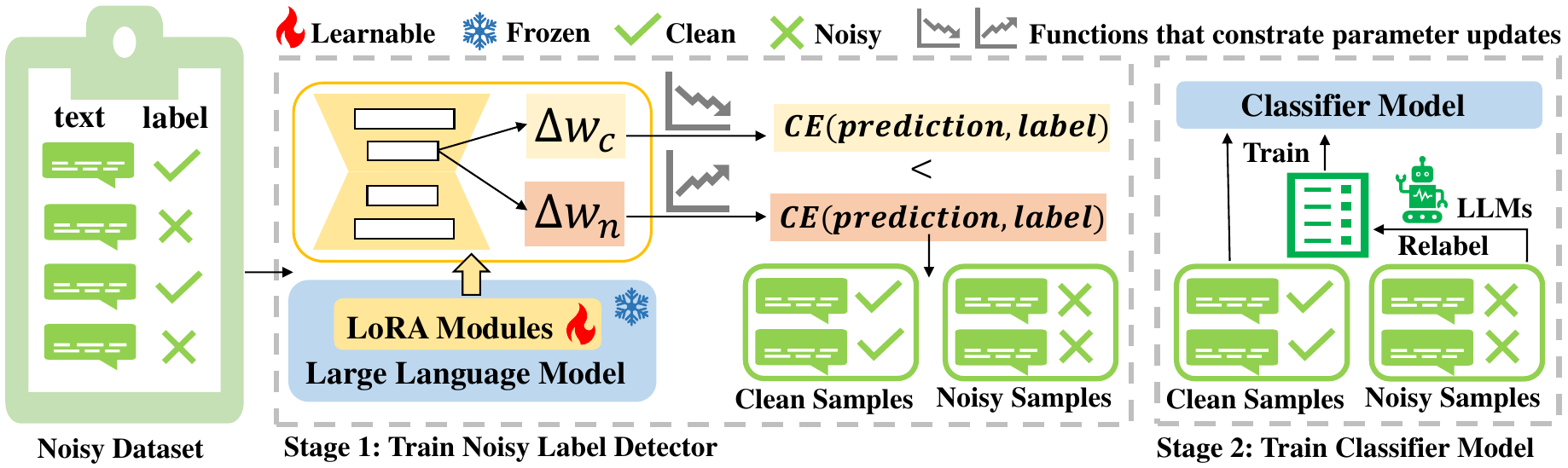}
\vspace{-0.2cm}
\caption{The architecture of our proposed framework Delora. Stage 1: We introduce two separate LoRAs (clean LoRA $\Delta w_c$ and noisy LoRA $\Delta w_n$) to construct the noisy label detector. Stage 2: We leverage the selected clean samples and relabeled noisy samples to train the classifier model.
}
\label{Framework}
\end{figure*}

\section{Method}
In this section, we present our proposed denoising learning framework Delora in detail. The main idea is to decouple sample selection and model training during the fine-tuning of LLMs on downstream tasks, avoiding the issue of vicious cycles. Overall, Delora comprises two pivotal stages. 
In the first stage, Delora introduces dual LoRAs to construct a noisy label detector, selecting clean samples and noisy samples. In the second stage, Delora leverages the curated clean samples and relabeled noisy samples to 
train the classifier model, further boosting the performance of our framework on text classification in the context of noisy environments. Figure \ref{Framework} shows our proposed framework.

\subsection{Identifying Noisy Labels with Dual LoRAs}
\label{Identifying Noisy Labels with Dual LoRAs}
In this stage, our core challenge lies in the construction of a noisy label detector for identifying noisy samples. The previous strategy is to set a fixed threshold
for loss value, such that the clean samples are associated with a smaller mean loss value and noisy
ones with bigger values. However, these methods \cite{DBLP:conf/coling/QiaoDDLCC022, DBLP:conf/mm/FengTP24} rely on the manual setting of thresholds, which restricts its practical applicability. Inspired by a recent study \cite{DBLP:journals/corr/abs-2412-04465}, we introduce two distinct LoRAs memorizing the information of clean and noisy samples to overcome these limitations.

\paragraph{Introducing dual LoRAs to construct a noisy label detector.} Given a language model with weights {$w^i_0$}, 
we introduce two LoRAs: clean LoRA $L_c = {\Delta w^i_c}$ and noisy LoRA $L_n = {\Delta w^i_n}$. Here, $i$ denotes the index of layers of transformers. For simplicity, we drop the superscript $i$ since our method operates over all the LoRA-enabled weight matrices of the base model. 
The parameters of $\Delta w_c$ are ideal parameters to memorize clean samples (desired memorization), while the parameters of $\Delta w_n$ are noisy parameters to memorize noisy samples (undesired memorization). During training, the clean LoRA $\Delta w_c$ aims to uncover distinguishable features by minimizing the cross-entropy (CE) between the predicted results of samples and their corresponding label, while the noisy LoRA $\Delta w_n$ serves as a learnable sample-dependent cross-entropy threshold to select clean samples. Specifically, the learnable threshold $\phi_i$ for the $i$-th training sample $x_i$ with a label $y_i$ is formulated as:
\begin{align}
\label{eq:threshold}
\phi_i = CE (f(x_i, w_0  + \Delta w_n), y_i),
\end{align}
which represents the cross-entropy between the observed label and the undesired prediction generated by noisy LoRA $\Delta w_n$. Based on this threshold, the clean subset $\mathcal{D}_c$ of $\mathcal{D}$ can be constructed as follows:
\begin{align}
\label{eq-phi}
\mathcal{D}_c = \left\{ (x_i, y_i) \mid  CE 
 (f(x_i, w_0  + \Delta w_c), y_i) < \phi_i \right\}.
\end{align}
The proposed selection strategy outperforms traditional small loss-based approaches in two ways: (1) it is more practical by using data-driven thresholds, which eliminates the need for manual setting; (2) the introduction of PEFT modules (LoRAs) improves its robustness to label noise, allowing it to identify challenging hard noise. However, the key challenge is optimizing dual LoRAs ($\textit{i.e.}$, the clean and noisy LoRA), and the noisy label detector. 


\paragraph{Optimization for Dual LoRAs.} Studies of the memory effect show that deep networks would first memorize clean samples and then noisy samples. From the perspective of network parameter updates, if we can strengthen the updates for parameters of clean LoRA $\Delta w_c$ during early training, and weaken the updates in later training, the clean samples can be better memorized by $\Delta w_c$. On the other hand, for noisy LoRA $\Delta w_n$, we should limit their parameter $\Delta w_n$ updates in early training, and strengthen their updates in later training to memorize noisy samples. This intuition drives us to formulate a new optimization objective for the dual LoRA to achieve our goal. 
To be specific, we design the corresponding optimization objective:
\begin{equation}
\begin{aligned}
\mathcal{L}_{LoRA}= \tau_1(t) \Delta \sigma_{c} +\tau_2(t) \Delta \sigma_{n},
\end{aligned}
\end{equation}
where $\tau_1(t)$ and $\tau_2(t)$ are two mathematical functions that are relevant to the training epoch $t$, $\Delta \sigma_c = ||\Delta w_c^t-\Delta w_c^{t-1}||=||\sigma_c^t(B)-\sigma _c^{t-1}(B)||+||\sigma_c^t(A)-\sigma_c^{t-1}(A)||$ is defined to measure the parameter change of clean LoRA $\Delta w_c$ between two adjacent epochs via the Euclidean distance, $\Delta w_c^t$ are the parameters of $\Delta w_c$ obtained in epoch $t$, $\sigma_c^t(A)$ and $\sigma_c^t(B)$ correspond to the parameters of two low-rank matrices in epoch $t$, respectively. More specifically, $\Delta \sigma_c$ limits the parameter change of $\Delta w_c$ between two adjacent epochs. 
If the weight, $\textit{i.e.}$, $\tau_1(t)$, is high, $\Delta \sigma_c$ will decrease quickly. Namely, modifications to $\Delta w_c^t$ are restricted.  
For noisy LoRA, the update of $\Delta \sigma_n  = ||\Delta w_n^t-\Delta w_n^{t-1}||=||\sigma_n^t(B)-\sigma _n^{t-1}(B)||+||\sigma_n^t(A)-\sigma_n^{t-1}(A)||$ is also constrained in a similar manner.

\textbf{Define for $\tau_1$ and $\tau_2$.} As analyzed, combining the memory effect of deep networks, we should 
dynamically adjust the parameter update of clean LoRA $\Delta w_c$ and noisy LoRA $\Delta w_n$. That is to say, during the early training, 
$\Delta w_c$ should be updated rapidly to fit clean data; while in later stages, its updates should slow down to prevent overfitting mislabeled data. In contrast, the update pattern for 
$\Delta w_n$ follows the opposite strategy. Therefore, we define $\tau_1(t)$ as a rising function to constrain the parameter update ($\Delta \sigma_c$) of clean LoRA $\Delta w_c$ and $\tau_2(t)$ as a decreasing function to constrain the parameter update ($\Delta \sigma_n$) of noisy LoRA $\Delta w_n$. 
In this work, we set $\tau_1(t) = t^{h_1}$ and $\tau_2(t) = t^{-h_2}$, where $h_1$ and $h_2$ are two hyperparameters. 
\paragraph{Optimization for the Noisy Label Detector.} 
After addressing the optimization challenges of the dual LoRA, we shift our focus to optimizing the noisy label detector. For the noisy label detector, optimizing the threshold $\phi_i$ in an end-to-end manner might be challenging due to its indirect involvement in forward propagation. We can still optimize the associated parameters by stating the clean probability of the $i$-th samples as follows:
\begin{equation}
\begin{aligned}
p_{i}^{\text{c}} = \frac{e^{CE (f(x_i, w_0  + \Delta w_c), y_i)}}
{e^{CE(f(x_i, w_0  + \Delta w_c), y_i)}+e^{CE (f(x_i, w_0  + \Delta w_n), y_i)}}
\end{aligned}
\end{equation}
Obviously, selecting samples with clean probability $p_i^c \textgreater 0.5$ corresponds to the criterion provided in Eq (\ref{eq-phi}). In this way, the original threshold-based selection method can be converted into binary classification problems for determining if a sample is clean or not. Specifically, given a sample $(x_i, y_i)$ from $\mathcal{D}$, we can use the dual LoRAs as a binary classifier to detect label noise. If the classifier makes a positive prediction of $x_i$, the sample is classified as clean.

To optimize the noisy label detector, we need to create positive and negative training samples for binary classification, \textit{i.e.}, correctly and incorrectly labeled samples. 
Inspired by the work \cite{DBLP:conf/iccv/KimYYK19} on negative learning,
for each text $x_i \in \mathcal{D}$, we randomly flip its class label $y_i$ to one of the other classes, \textit{i.e.}, $y^n_i \in \left\{1,\dots,K\right\}$ \textbackslash $\left\{y_i\right\}$, to construct negative datasets $D_n$. 
For positive datasets, we first use LLMs (\textit{i.e.}, GPT-4o) to generate pseudo-labels for each text $x_i \in \mathcal{D}$, then select samples where the generated pseudo-labels match the original labels $y_i$ to construct the positive datasets $D_p$. Here, we treat the samples in $D_p$ as positive samples and the samples in $D_n$ as negative samples (see Appendix \ref{negative dataset construction} for details).
After that, 
the optimization objective can be calculated as follows:
\begin{equation}
\begin{aligned}
\mathcal{L}_{Detector} = \frac{1}{N_p} \sum_{i=1}^{N_p} \ell_{nll} (p_{i}^{c}, y_i) + \ell_{nll} (1 - p_{i}^{c}, y_i^n),
\end{aligned}
\end{equation}
where $N_p$ denotes the size of $D_p$, $D_p \subseteq D$,
\( \ell_{nll}(\cdot, \cdot) \) is the negative log-likelihood loss, defined as  
$\ell_{nll}(p_i, y) = -\log p_{iy}$.

Although the design of $\mathcal{L}_{LoRA}$ and $\mathcal{L}_{Detector}$ aim to help the noisy label detector perform sample selection, they focus primarily on the comparison between different LoRAs (\textit{i.e.}, $\Delta w_c$ and $\Delta w_n$). To enhance the noisy label detector's ability to learn task-specific representations, we introduce the cross-entropy loss 
$\mathcal{L}_{ce}=-\frac{1}{N}\sum_{i=1}^{N} \mathrm{log}f(x_i, w_0  + \Delta w_c+\Delta w_n)$. It is worth noting that the optimization objective $\mathcal{L}_{warm} = \mathcal{L}_{ce} + \mathcal{L}_{LoRA}$ is computed in the warm-up stage, the optimization objective $\mathcal{L} = \mathcal{L}_{ce}+\mathcal{L}_{LoRA} + \mathcal{L}_{Detector}$ is computed after warm-up.
With these objectives, Delora effectively identifies noisy samples using learned thresholds.




\subsection{Training Classifier Model after Selection}
After obtaining the selected clean and noisy samples, we introduce the classifier model training stage to learn a classifier model using selected samples. For selected clean samples, we directly utilize the cross-entropy loss to learn from them.
For selected noisy samples, we refine and leverage them following the recent work \cite{DBLP:conf/acl/YuanCZJ24}, rather than discarding them as done in most previous works \cite{DBLP:conf/coling/QiaoDDLCC022, DBLP:conf/acl/KimK024}. Specifically, we leverage clean samples to construct demonstrations prompting GPT-4o to relabel noisy samples. Then, we utilize the robust loss function to learn from the relabeled noisy samples. See Appendix \ref{Appendix: Loss functions} for more details. Notably, the clean and noisy samples obtained through Eq. (\ref{eq-phi}) in the last stage, which enables the second stage to be broadly applicable to various classifier models (pre-trained language models or open-source LLMs) and fine-tuning paradigms (full fine-tuning or PEFT), regardless of their backbone architectures. We validate the versatility of Delora in Section \ref{Analysis}.

\section{Experiments}
\begin{table*}[htb!]
    \centering
    \small
    \setlength{\tabcolsep}{2.7pt} 
    \begin{tabular}{cc|cccc|cccc|cccc}
        \toprule
        \multirow{2}{*}{\textbf{Dataset}}&
        \multirow{2}{*}{\textbf{Method}} & \multicolumn{2}{c}{20\%S} & \multicolumn{2}{c|}{40\%S} & \multicolumn{2}{c}{20\%A} & \multicolumn{2}{c|}{40\%A} & \multicolumn{2}{c}{20\%I} & \multicolumn{2}{c}{40\%I} \\
        \cmidrule(lr){3-6} \cmidrule(lr){7-10}
        \cmidrule(lr){11-14}
        & & Prec. & Rec. & Prec. & Rec. & Prec. & Rec. & Prec. & Rec. & Prec. & Rec. & Prec. & Rec. \\
        \midrule
        \multirow{3}{*}{\textbf{Trec}} & LLMs-detection & 70.37 & 70.31 & 70.16 & 70.20 & 69.96 & 69.97& 69.04 & 68.82 &-&-&-&- \\        
        & Small-loss & 81.15 & 88.55 & 60.16 & 87.82 & 81.53 & 74.02& 59.41 & 96.20 &-&-&-&- \\
        & Delora (Ours) & \colorbox{green!20}{\textbf{99.47}}&\colorbox{green!20}{\textbf{95.30}} & \colorbox{green!20}{\textbf{99.28}} & \colorbox{green!20}{\textbf{96.44}} &  \colorbox{green!20}{\textbf{99.19}} & \colorbox{green!20}{\textbf{98.06}} & \colorbox{green!20}{\textbf{99.12}} & \colorbox{green!20}{\textbf{97.27}} &-&-&-&- \\   
        \midrule
        \multirow{3}{*}{\textbf{SST-5}} & LLMs-detection & 60.91 & 60.71 & 59.93&59.92 & 59.80 & 59.85 & 59.46 & 59.38 & 67.11& 65.47 & 65.56&67.58\\
        & Small-loss & 80.83&79.35&59.83&78.45&79.83&77.85&59.26&76.63&80.01&74.03&60.11&73.96\\
      \rowcolor{green!20}  \cellcolor{white}{}  & \cellcolor{white}{ Delora (Ours)}  &
 {\textbf{98.11}}&  {\textbf{86.75}}& {\textbf{96.37}}& {\textbf{93.59}}&\textbf{97.18}&\textbf{94.90}&\textbf{94.04}&\textbf{92.61}&\textbf{95.31}&\textbf{85.02}&\textbf{91.99}&\textbf{89.29} \\  
        \midrule
        \multirow{3}{*}{\textbf{SST-2}} & LLMs-detection & 78.94 & 79.62 & 79.41 & 78.78 & 79.53 & 79.31 & 77.88 & 77.53 & 85.97 & 85.08 & 84.58 & 85.64\\
        & Small-loss & 80.87&  85.78&  59.75& 80.80 & 80.07 & 85.48 & 61.20 & 68.34 & 80.59 & 74.95  & 61.27 & 81.78\\
     \rowcolor{green!20}  \cellcolor{white}{}  & \cellcolor{white}{ Delora (Ours)} &  \textbf{99.97}&\textbf{89.07}&\textbf{99.95}&\textbf{86.34}&\textbf{99.96}&\textbf{88.60}&\textbf{99.76}&\textbf{86.75}&\textbf{98.35}&\textbf{87.27}&\textbf{96.43}&\textbf{88.86}\\
        \bottomrule[0.1pt]
        
    \end{tabular}
    \vspace{-0.2cm}
    \caption{We compare the Precision (\%) and Recall (\%) of Delora with LLMs-detection and small-loss to evaluate the 
    performance of noisy label detection (\textit{i.e.}, clean sample selection performance). \textbf{Bold} means the best score.}    
    \label{results: noisy label detection}
\end{table*}

\begin{table*}[htb!]
    \footnotesize
     \caption{Performance (test accuracy \%) comparison of Delora with other LNL baselines on synthetic noise datasets. Base (Clean) refers to the base model trained on ground truth data without noisy labels. LLM-base refers to directly using LLMs (GPT-4o)
      on the test dataset. \textbf{Bold} means the best score for each dataset.}
    \label{Table: Main results}
    \centering
   \renewcommand\arraystretch{1.5}
\setlength\tabcolsep{1.6pt}
    \begin{tabular}{c |c c c c|c c c c c c |c c c c c c}
    \toprule
  
\multirow{2}*{\textbf{Model}}   &\multicolumn{4}{c|}{\textbf{ Trec}} &
    \multicolumn{6}{c|}{\textbf{SST-5}}  & \multicolumn{6}{c}{\textbf{SST-2}}  \\  
    \cmidrule(lr){2-5} \cmidrule(lr){6-11}\cmidrule(lr){12-17}
  & 20\%\textbf{S} & 40\%\textbf{S} & 20\%\textbf{A} & 40\%\textbf{A} & 20\%\textbf{S} & 40\%\textbf{S} & 20\%\textbf{A} & 40\%\textbf{A} & 20\%\textbf{I} & 40\%\textbf{I} & 20\%\textbf{S} & 40\%\textbf{S} & 20\%\textbf{A} & 40\%\textbf{A} & 20\%\textbf{I} & 40\%\textbf{I}\\
  \midrule
  Base (Clean)  & \multicolumn{4}{c|}{98.60} 
& \multicolumn{6}{c|}{58.05}
& \multicolumn{6}{c}{97.03}\\
  Base  & 95.20 & 90.20 & 94.20 &  87.40 & 
  54.08 & 49.59& 54.81& 47.70& 53.07& 46.76 & 86.43 & 64.62 &
  86.70 & 65.88 & 83.85  & 63.15\\
  \midrule
  LLM-base  & \multicolumn{4}{c|}{71.35} & \multicolumn{6}{c|}{70.51} & \multicolumn{6}{c}{91.51} \\
  \midrule
  Co-Teaching &  95.51 & 90.98 & 95.32 & 89.24 & 53.99 & 49.72 & 55.07 & 47.24 & 52.63 & 46.45 & 87.29 & 67.21 & 89.69 & 69.60 & 85.59 & 67.16\\
\midrule
SENT &  95.49 & 91.25 & 95.43 & 90.53 & 54.05 & 49.61 & 55.17 & 47.68 & 53.70 & 46.94 & 87.46 & 67.17 & 89.12 & 69.10 & 85.38 & 66.23\\
\midrule
   LAFT & 95.42 & 91.28 & 94.43 & 90.32 & 55.00 & 49.13 & 54.50 & 47.69 & 52.57 & 47.02 & 88.07 & 67.39 & 89.72 & 68.80 & 85.45 & 66.34\\
  \midrule  
   SelfMix & 96.21 & 90.52 & 95.24 & 90.80  & 53.63 & 49.61 & 55.80 & 47.59 & 53.00 & 46.66 & 87.58 & 66.78 & 89.97 & 68.92 & 85.61 & 66.05\\
  \midrule
   CleaR & 96.01 & 90.45 & 95.35 & 90.69 & 53.69 & 49.97 & 54.95 & 47.63 & 53.64 & 46.62 & 87.21 & 66.81 & 89.36 & 69.43 & 85.13 & 66.54\\
  \midrule
   NoiseAL & 97.30 & 96.54 & 96.96 & 95.95 & 55.00 & 50.48 & 54.94 & 48.12 & 54.00 & 47.32 & 91.90 & 86.25 & 91.97 & 86.72 & 91.55 & 85.01\\
  \midrule
 \rowcolor{green!20}   \cellcolor{white}{ Delora (Ours)}&\textbf{98.46} & \textbf{97.60} & \textbf{98.30}& \textbf{97.40}& \textbf{57.39}& \textbf{55.62}& \textbf{57.57}& \textbf{55.39}& \textbf{57.02}& \textbf{55.02}& \textbf{96.50} &\textbf{95.75}&\textbf{96.27}&\textbf{95.18}&\textbf{96.08}&\textbf{95.00}\\
  \midrule
    \end{tabular}
      \vspace{-0.3cm}
\end{table*}

\subsection{Experimental Settings}
\paragraph{Synthetic Datasets.} We first fully evaluate our approach on five text classification benchmarks by synthesizing noisy labels with a variety of noise types and ratios. Specifically, the experiments are evaluated on the following benchmark datasets: Trec \cite{DBLP:conf/coling/LiR02}, SST-2 \cite{DBLP:conf/emnlp/SocherPWCMNP13}, SST-5 \cite{DBLP:conf/emnlp/SocherPWCMNP13}, 
20ng \cite{DBLP:conf/icml/Lang95}, AGNews \cite{DBLP:conf/www/Gulli05}. Following the previous experimental setup \cite{DBLP:conf/coling/QiaoDDLCC022, DBLP:conf/acl/YuanCZJ24}, we then artificially introduce the noise by using three different strategies: (1) \textbf{Symmetric Noise (S)} chooses one of the other classes at random to replace the label;
(2) \textbf{Asymmetric Noise (A)} carries out pairwise label flipping, in which a class $i$ can only change to the following class ($i$ mode $K$) + 1;
(3) \textbf{Instance-dependent Noise (I)}
alters labels according to the transition probability determined by the related attributes of the instance. For all kinds of noise, the noise rate is set to 20\% and 40\%. See Appendix \ref{Appendix: Detailed Process for Generating Noisy Labels} for more details.

\textbf{Real-World Datasets.} We further evaluate our approach on three real-world datasets with noisy labels: 
Yorùbá\cite{DBLP:conf/emnlp/HedderichAZAMK20},
Hausa \cite{DBLP:conf/emnlp/HedderichAZAMK20}, 
AlleNoise \cite{DBLP:journals/corr/abs-2407-10992}. See Appendix \ref{Appendix: Details of Real-World Datasets with Noisy Labels} for more details.

\textbf{Baselines for Noisy Label Detection.}
To justify the effectiveness of Delora in detecting noisy labels, we first compare our approach with two other sample selection strategies: (1) \textbf{LLMs-detection} strategy, which directly determines whether a given sample is clean using the LLM (\textit{i.e.}, GPT-4o).
(2) \textbf{Small-loss} strategy, which selects some samples with small training losses as clean via the Gaussian Mixture Model, 
appearing in most LNL works.

\textbf{Baselines for LNL.} Then, we further compare Delora with previous sample selection-based LNL baselines as follows: (1) Base model (Llama3.1-8B-Instruct \cite{DBLP:journals/corr/abs-2407-21783}) without noise-handling;
(2) Some methods that use small-loss strategy:
\textbf{Co-Teaching} \cite{DBLP:conf/nips/HanYYNXHTS18},
\textbf{SelfMix} \cite{DBLP:conf/coling/QiaoDDLCC022}, 
\textbf{NoiseAL} \cite{DBLP:conf/acl/YuanCZJ24},
\textbf{CleaR} 
\cite{DBLP:conf/acl/KimK024}; (3) Other technologies : \textbf{SENT}\cite{DBLP:conf/emnlp/WangLWCLZCY22},
\textbf{LAFT} \cite{DBLP:conf/emnlp/WangTGL23}. 
See Appendix \ref{Appendix: Details of Baselines} for more details.

\textbf{Evaluation Metrics.} For the first stage (noisy label detection stage), we evaluate the selection of clean samples using precision and recall metrics.  A higher recall suggests that more clean samples are found in the noisy dataset, whereas a higher precision means that there are more real clean examples in $D_c$. For the last stage (classifier model training stage), we use the test accuracy to evaluate the generalization performance of our framework on text classification in the context of noisy environments.

The implementation details are in Appendix \ref{Appendix: Implement Details}.

\subsection{Performance for Noisy Label Detection}
Table \ref{results: noisy label detection} presents the performance for noisy label detection on three synthetic datasets. The results presented in Table \ref{results: noisy label detection} show that our proposed noisy label detector outperforms the comparison strategies across all dataset settings, demonstrating significant improvements in precision and recall. The LLMs-detection strategy directly leverages the zero-shot ability of LLMs to perform binary classification (clean or noisy). The lack of targeted demonstrations in LLM prompts has resulted in lower selection performance for this strategy. In contrast, the small-loss strategy and Delora make better use of training samples to perform better in precision and recall. Moreover, leveraging the different LoRA modules to memorize the information of clean and noisy samples separately, Delora surpasses the small-loss strategy in detecting noisy labels, particularly under severe noise settings and fine-grained classification datasets. For instance, in datasets with a noise rate of 40\%, the small-loss strategy performs poorly in precision, while our method performs well in various situations. Additionally, Delora reduces the need to manually set loss thresholds, making it a practical and effective approach to detect label noise in real-world tasks.
\begin{table}[th]
 \setlength\tabcolsep{8pt}
\centering
\scalebox{0.8}
{
\begin{tabular}{l c c c}
\toprule[1.3pt]
\textbf{Method}  & \textbf{Hausa} & \textbf{Yorùbá} & \textbf{AlleNoise}\\
\midrule[0.9pt]
Noise Ratio & 50.37\% &33.28\% &15.00\% \\
\midrule[0.9pt]
Base &  49.80$\pm _{±0.26}$ & 67.02$\pm _{±0.32}$ & 65.75$\pm _{±0.25}$\\
Co-Teaching & 46.47$\pm _{±0.38}$ & 66.23$\pm _{±0.69}$ & 65.32$\pm _{±0.36}$\\
SENT & 46.15$\pm _{±0.57}$ & 66.21$\pm _{±0.24}$ & 66.66$\pm _{±0.38}$\\
LAFT & 50.56$\pm _{±0.31}$ & 69.13$\pm _{±0.54}$ & 66.57$\pm _{±0.81}$\\
SelfMix & 50.81$\pm _{±2.62}$ & 69.27$\pm _{±1.07}$ & 67.67$\pm _{±3.67}$\\
CleaR & 51.66$\pm _{±1.14}$ & 70.11$\pm _{±1.10}$ & 67.73$\pm _{±2.40}$\\
NoiseAL & 52.34$\pm _{±0.69}$ & 72.13$\pm _{±0.24}$ & 69.80$\pm _{±0.62}$\\
\rowcolor{green!20}  \cellcolor{white}{ Ours}& \textbf{60.12}$\pm _{±0.22}$ & \textbf{78.56}$\pm _{±0.30}$ & \textbf{76.28}$\pm _{±0.23}$\\
\bottomrule[1.3pt]
\end{tabular}
}
\caption{Main results on real-world noise datasets}
\vspace{-0.2cm}
\label{Table: real-world noise}
\end{table}

\subsection{Performance for LNL}
Table \ref{Table: Main results} and Table \ref{Table: real-world noise}
show the main results for three synthetic and real-world noisy datasets (more results in Appendix \ref{Appendix: More detailed Results}). From these results, we found that (1) Delora significantly outperforms all baselines on synthetic datasets with varying noise types and ratios, which validates the effectiveness of our proposed two-stage decoupling framework in addressing the LNL task. (2) For fine-grained classification datasets, while most compared methods show little improvement on SST-5 due to its fine-grained nature, Delora shows substantial improvement. (3) For real-world datasets with a high noise rate, previous methods show limited improvement on Hausa. In contrast, Delora achieves a significant performance boost, showcasing its ability to combat label noise in practical situations.

\subsection{Ablation Studies}
\label{Ablation Studies}
We perform ablation studies to investigate the contributions of each component and routing strategy in Delora. Table \ref{Table:Ablation} presents the ablation results.

\textbf{The effect of noisy label detector.}
Noisy Label Detector (NLD) effectively identifies noisy samples and partitions the noisy dataset, which can alleviate the overfitting of noisy labels and the issue of vicious cycles in the subsequent classifier model training stage. The performance of Delora will decrease greatly when we remove the NLD, which indicates that utilizing the NLD is crucial. 

\textbf{The effect of classifier model training.} The classifier model training stage (CT) leverages the selected samples to fine-tune the classifier model, obtaining the best classification performance. If we remove CT and only rely on the trained clean LoRA to perform classification tasks, the performance of Delora will decrease by a large margin, which 
emphasizes the necessity of CT in Delora.

\textbf{The effect of different optimization objectives.} In our experiments, we combine different optimization objectives ($\mathcal{L}_{LoRA}, \mathcal{L}_{Detector}, \mathcal{L}_{ce}$) to train the noisy label detector. The results in Table \ref{Table:Ablation} demonstrate that each component is essential for improving generalization and robustness, failing to employ them leads to a decline in the results.

\textbf{The effect of selected noisy samples.} To maximize the use of training data without discarding remaining noisy samples (NS), we relabel and repurpose them to further train the classifier model. This process has also been proven to be crucial, as its removal leads to substantial performance drops.

\begin{table}[h]
\setlength\tabcolsep{5pt}
    \centering
    \begin{tabular}{l|cc|cc}
        \toprule
        \multirow{2}{*}{\textbf{Variant}} & \multicolumn{2}{c|}{\textbf{Sym}} & \multicolumn{2}{c}{\textbf{Asym}} \\
        & 20\% & 40\% &  20\% & 40\% \\
        \midrule
        Delora (Ours) & \textbf{98.46} & \textbf{97.60} & \textbf{98.30}& \textbf{97.40} \\
        \midrule
        w/o NLD & 95.20 & 90.20 & 94.20 & 89.40\\
        w/o CT & 96.02 & 91.01 & 94.77 & 90.08\\
        w/o $\mathcal{L}_{LoRA}$ & 96.46 & 92.00 & 95.68 & 90.43 \\
        w/o $\mathcal{L}_{Detector}$& 96.91 & 91.98 & 95.07 & 90.75 \\
        w/o $\mathcal{L}_{ce}$& 96.54 & 91.33 & 95.44 & 90.35 \\
        w/o NS & 97.41 & 94.29 & 96.97 & 91.53\\ 
        \bottomrule
    \end{tabular}
    \caption{Ablation study on the Trec dataset.}
    \vspace{-0.2cm}
    \label{Table:Ablation}
\end{table}

\vspace{-0.2cm}
\subsection{Analysis}
\label{Analysis}
\begin{figure}[htbp]
  \centering
  \begin{minipage}[t]{1\linewidth}  
      \centering
    \label{fig:subfig5}\includegraphics[width=7.8cm, height=3.5cm]{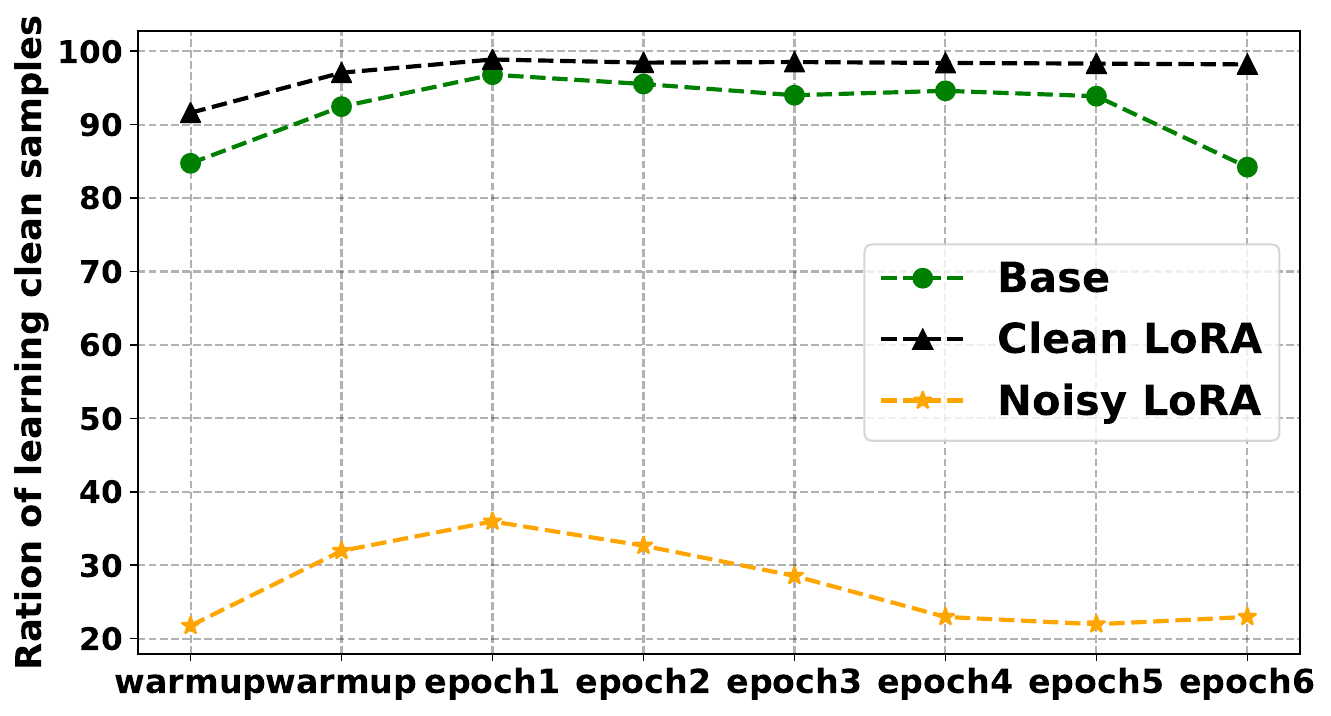}
      \subfigure{(a) Impact on learning clean samples}
  \end{minipage}
  \begin{minipage}[t]{1\linewidth}
      \centering
      \label{fig:subfig6}\includegraphics[width=7.8cm, height=3.5cm]{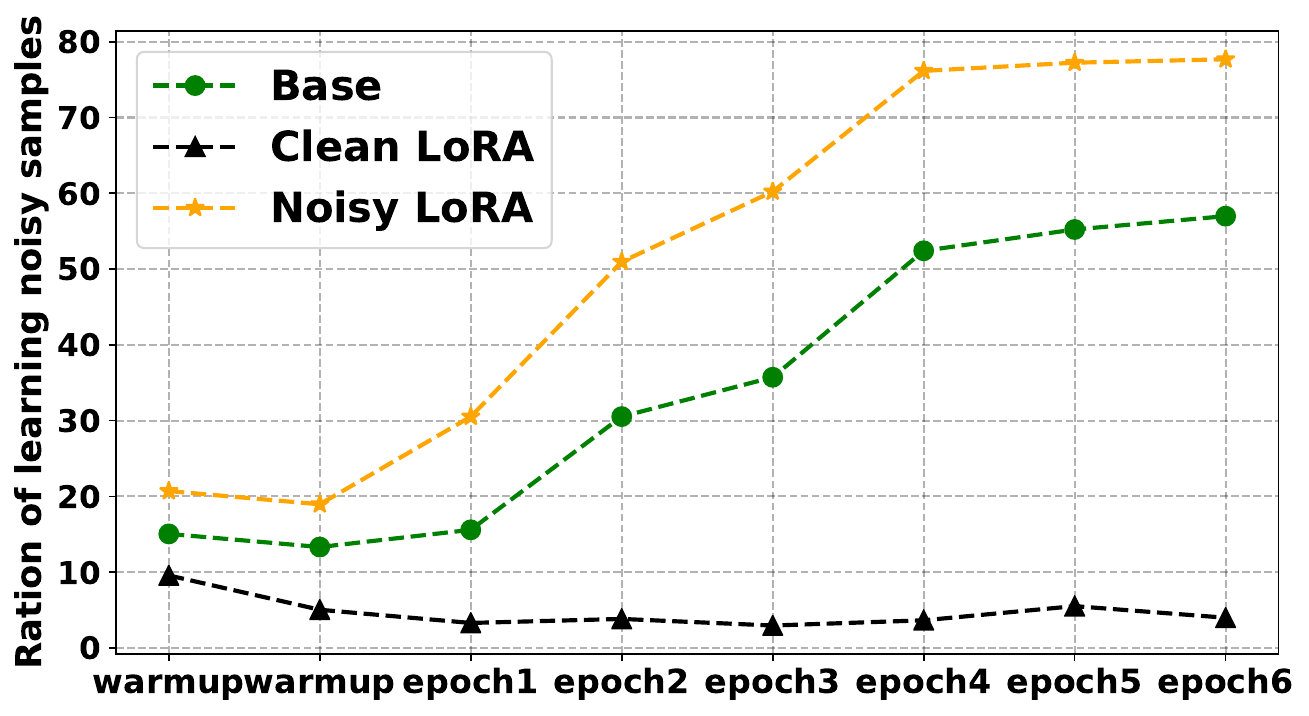}
      \subfigure{(b) Impact on learning noisy samples}
  \end{minipage}
  \vspace{-0.3cm}
  \caption{Memorization performance of different LoRAs during fine-tuning on Trec under 20\%A. The green line refers to the base model without noise handling.}
  \label{Memorization performance of different LoRAs.}
\end{figure}

\textbf{Memorization performance for different LoRAs.} In our proposed Delora, we adopt the clean and noisy LoRA to memorize the information of clean and noisy samples separately. To confirm whether different LoRA modules fit clean and noisy samples, we compare the ratio of memorizing clean and noisy samples for different LoRAs during fine-tuning. As shown in Figure \ref{Memorization performance of different LoRAs.}, we observe that the base model first memorizes clean samples and then gradually fits noisy samples, which is consistent with the memory effect. For clean LoRA, it enhances the memorization of clean samples while reducing the memorization of noisy samples. In contrast, noisy LoRA exhibits the opposite effect, the parameters in noisy LoRA are restricted to absorb the side effects of mislabeled data. These results indicate that Delora effectively constrains different LoRA modules to memorize clean and noisy samples separately, thereby facilitating the successful construction of the noisy label detector.

\begin{figure}[h!]
    \vspace{-0.2cm}
  \centering
    \subfigure[Precision On Trec]{\includegraphics[width=0.49\textwidth,height=0.35\textwidth]{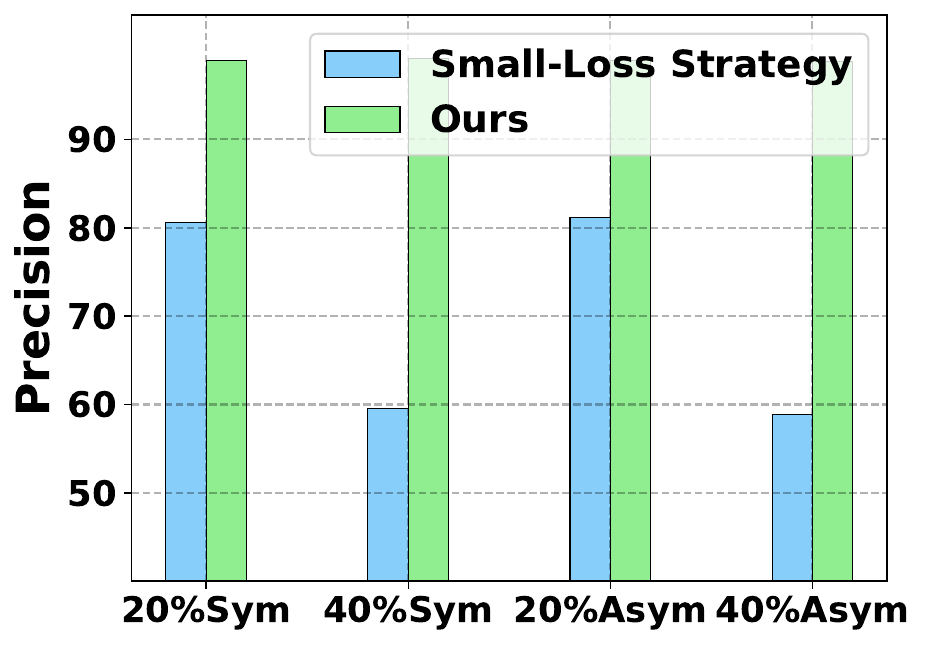}}
    \subfigure[Recall On Trec]{\includegraphics[width=0.49\textwidth,height=0.35\textwidth]{ 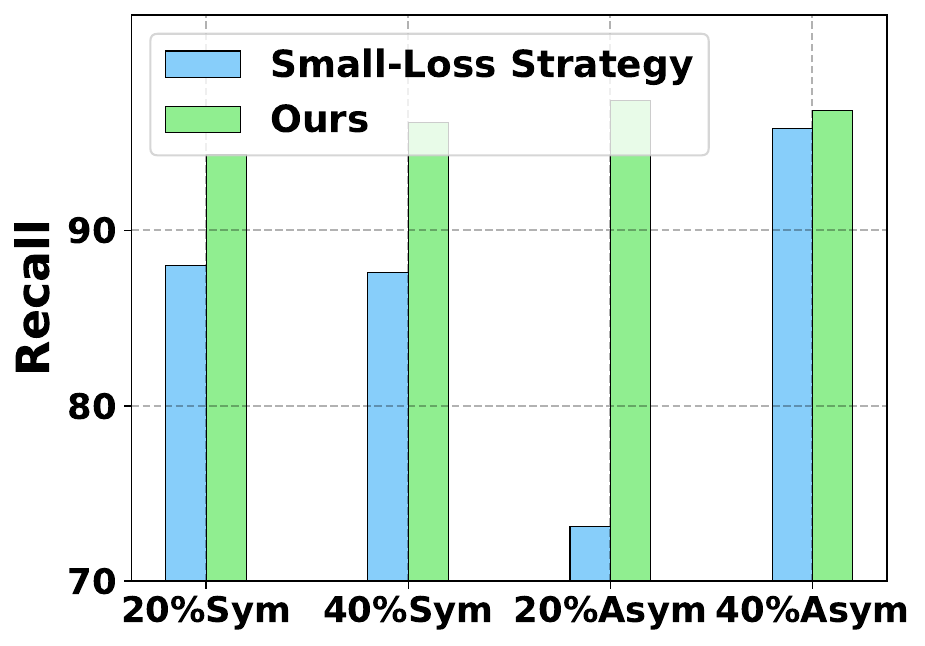}} \\
        \vspace{-0.3cm}
  \caption{Precision and recall of noisy label detection with BERT as the backbone on the Trec dataset.}
    \label{Precision and recall of noisy label detection with BERT as the backbone.}
\end{figure}

\textbf{Impact of Noisy Label Detector Backbones.} 
Table \ref{results: noisy label detection} has demonstrated the superiority of Delora in detecting noisy labels based on Llama 3.1-8B. To evaluate the impact of different language model backbones on noise detection, we use the small language model BERT as the backbone and report the results in Figure \ref{Precision and recall of noisy label detection with BERT as the backbone.}. As shown in Figure \ref{Precision and recall of noisy label detection with BERT as the backbone.}, our method maintains consistently superior performance, underscoring its resilience across different backbones. We analyze how our method consistently outperforms others across different language models. Generally, for LNL, incorrect labels can propagate errors during backpropagation, causing adjustment bias across all trainable parameters. Delora effectively mitigates this issue by ensuring that the influence of incorrect labels is restricted to the noise-specific parameters within the noise LoRA, thereby preserving the integrity of the clean LoRA, which is dedicated to learning from clean data. This solution achieves functional segregation at the parameter level, which is architecture-independent.

\textbf{Delora for Various Classifier Models.} The results in Table \ref{Table:Ablation} highlight the crucial role of the classifier model training stage in our method. Notably, Delora seamlessly integrates with various classifier models at this stage. To showcase the versatility of our proposed LNL framework, we conduct experiments on Trec using a diverse set of models, including BERT (full fine-tuning), Llama 
 3.2-3B \cite{DBLP:journals/corr/abs-2407-21783}, 
Gemma 2-9B \cite{DBLP:journals/corr/abs-2408-00118}, Llama 3.1-8B. As shown in Table \ref{Table: comparison with various classifier models}, Delora exhibits the best performance on average.

\begin{table}[th]
 \setlength\tabcolsep{5pt}
\centering
\scalebox{0.8}
{
\begin{tabular}{l c c c|c}
\toprule[1.3pt]
\textbf{Architecture}  & \textbf{BERT} & \textbf{Llama3.2} & \textbf{Gemma2} & \textbf{Llama3.1}\\
\midrule[0.9pt]
Base &  93.60 & 93.88&  94.49 & 94.20\\
Co-Teaching & 94.88 &  94.81 &  95.74 & 95.32 \\
SelfMix & 95.16 & 96.05 & 95.12 & 95.24\\
NoiseAL & 96.80 & 96.23 & 97.14 & 96.96\\
Delora & \textbf{97.40} & \textbf{97.50} & \textbf{98.12} & \textbf{98.30}\\
\bottomrule[1.3pt]
\end{tabular}
}
\caption{Test accuracy (\%) using various classifier models on Trec under 20\%A. \textbf{Bold} means the best score.}
\vspace{-0.2cm}
\label{Table: comparison with various classifier models}
\end{table}
\vspace{-0.3cm}

\begin{table*}[htb!]
    \footnotesize
     \caption{Detailed results for Trec datasets with the LLaMA 3.1 8B as backbone.}
    \label{Table: Detailed results for Trec datasets with the LLaMA 3.1 8B as backbone}
    \centering
    \setlength\tabcolsep{4.5pt}
    \begin{tabular}{c | c c c c c c}
    \toprule
  Methods & Parameter (M) & Memory (GB) & 20\%S & 40\%S & 20\%A & 40\%A \\
  \midrule
  standard single LoRA &13.7&10.4&95.20&90.20 &94.20&87.40\\
  \midrule
  SelfMix &28.3&19.3&96.21 (+1.01)& 90.52 (+0.32)& 95.24 (+1.04) & 90.80 (+3.40)\\
  \midrule
  NoiseAL &28.4 & 21.7 & 97.30 (+2.10) & 96.54 (+6.34) & 96.96 (+2.76) & 95.95 (+8.55)\\
  \midrule
  \rowcolor{green!20}  \cellcolor{white}{ Ours}
& 27.3&13.6&98.46 (+3.26)&97.60 (+7.40)&98.30 (+4.10)&97.40 (+10.00)\\ 
    \bottomrule
    \end{tabular}
\end{table*}

\begin{table*}[htb!]
    \footnotesize
     \caption{Parameter efficiency on Trec datasets with different noisy settings.}
    \label{Table: parameter efficiency}
    \centering
    \setlength\tabcolsep{2.5pt}
    \begin{tabular}{c | c c c c}
    \toprule
  Methods & 20\%S & 40\%S & 20\%A & 40\%A \\
  \midrule
  Single LoRA $\rightarrow$ SelfMix & +0.07\%‌ / +1M params & +0.02\%‌ / +1M params & +0.07\%‌ / +1M params &  +0.23\%‌ / +1M params\\
  \midrule
    Single LoRA $\rightarrow$ NoiseAL & +0.19\%‌ / +1M params & +0.56\%‌ / +1M params & +0.24\%‌ / +1M params &  +0.76\%‌ / +1M params\\
  \midrule
\rowcolor{green!20}  \cellcolor{white}{ Single LoRA $\rightarrow$ Ours}   &   +0.24\%‌ / +1M params & +0.54\%‌ / +1M params & +0.30\%‌ / +1M params &  +0.74\%‌ / +1M params\\  
    \bottomrule
    \end{tabular}
\end{table*}

\begin{table*}[htb!]
    \footnotesize
     \caption{Memory efficiency on Trec datasets with different noisy settings.}
    \label{Table: memory efficiency}
    \centering
    \setlength\tabcolsep{2pt}
    \begin{tabular}{c | c c c c}
    \toprule
  Methods & 20\%S & 40\%S & 20\%A & 40\%A \\
  \midrule
  Single LoRA $\rightarrow$ SelfMix & +0.11\%‌ / +1G memory & +0.04\%‌ / +1G memory & +0.12\%‌ / +1G memory & +0.38\%‌ / +1G memory\\
  \midrule
    Single LoRA $\rightarrow$ NoiseAL & +0.14\%‌ / +1G memory &+0.43\%‌ / +1G memory & +0.19\%‌ / 1G memory & +0.58\%‌ / +1G memory\\
  \midrule

  \rowcolor{green!20}  \cellcolor{white}{Single LoRA $\rightarrow$ Ours}
 &   +1.02\%‌ / +1G memory &+2.31\%‌ / +1G memory&+1.28\%‌ / +1G memory &+3.13\%‌ / +1G memory
\\  
    \bottomrule
    \end{tabular}
\end{table*}

\textbf{Analysis of Efficiency.}
In this part, we explore the efficiency of our proposed methods. Although the proposed methods introduce additional trainable parameters, we perform the systematic multi-dimensional efficiency comparisons (see the  Table \ref{Table: Detailed results for Trec datasets with the LLaMA 3.1 8B as backbone}, from the perspective of final classification performance) to demonstrate that our method achieves a superior Pareto frontier in the accuracy-parameters-memory trade-off space. Specifically, from the Table \ref{Table: Detailed results for Trec datasets with the LLaMA 3.1 8B as backbone}, there are some key findings as follows: 

(1) Compared to the base model with the standard single LoRA, we require only ‌+13.6 M‌B parameters and ‌+3.2 GB‌ memory but achieve ‌+3.26\%‌, +7.40\%‌, +4.10\%‌, and +10\%‌ accuracy on the Trec dataset with different noisy settings, respectively.

(2) Compared to other baselines (SelfMix and NoiseAL), our proposed method outperforms them with relatively fewer resources. Moreover, we further compare our method with other baselines by quantifying parameter efficiency via ‌accuracy gain per parameter ($\Delta$Acc/$\Delta$Params)‌ (see the Table \ref{Table: parameter efficiency}) and memory efficiency via ‌accuracy gain per parameter ($\Delta$Acc/$\Delta$Memory) (see the Table \ref{Table: memory efficiency}).‌ In the above experiments, we chose the LLaMA 3.1 8B as the backbone. 

In general, while the dual LoRA modules introduce additional parameters, our systematic analysis of parameter efficiency ($\Delta$Acc/$\Delta$Params) and memory efficiency ($\Delta$Acc/$\Delta$Memory) demonstrates that the increased parameterization is justified. Moreover, through comparative analysis across different methods, we quantify the accuracy gain per additional 1M parameters or 1GB memory, ultimately proving that our approach achieves a superior Pareto frontier in the three-dimensional trade-off space of accuracy, parameter count, and memory usage.

\section{Conclusion}
In this work, we mitigate the issues of the vicious cycle in current mainstream sample selection methods and further explore the PEFT methods in the era of LLMs to solve the LNL tasks. Specifically, we decouple this task into sample selection and classifier model training. For sample selection, we introduce the dual LoRA to construct a new noisy label detection approach, which restricts the clean LoRA to fit clean data and the noisy LoRA to absorb the side effects of mislabeled data. Then we can train the robust classifier model by leveraging the carefully selected samples. Extensive experiments have convincingly demonstrated the superior performance of Delora in both noisy label detection and text classification. Moreover, our in-depth analysis has demonstrated that Delora can generalize to other language models.

\section*{Limitations}
While Delora has demonstrated significant improvements in text classification tasks, there are some limitations to consider in the following aspects:
(1) Due to resource constraints, we have not evaluated our framework on larger language models, such as Llama-3.2 70B.
(2) The experiments in this paper are limited to the text classification task and do not explore other tasks \cite{DBLP:journals/apin/WangZH22}, such as text generation tasks. Interestingly, we discovered a recent study \cite{luo2024robustftrobustsupervisedfinetuning} that extends the NoiseAL method, originally designed for text classification, to text generation tasks. This opens up new perspectives and potential improvements for our work. We leave the exploration of this direction as promising future work.




\clearpage
\appendix

\section{Construction of Positive and Negative Sample}
\label{negative dataset construction}
Given the original dataset $\mathcal{D}$, for each text $x_i \in \mathcal{D}$, we randomly flip its class label $y_i$ to one of the other classes, \textit{i.e.}, $y^n_i \in \left\{1,\dots,K\right\}$ \textbackslash $\left\{y_i\right\}$, to construct negative datasets $D_n$. For texts and their corresponding labels in $D_n$, we regard them as negative samples, which train the noisy label detector in such a logical form: "input text does not belong to this complementary label."

For positive datasets, we first use LLMs (\textit{i.e.}, GPT-4o) to generate pseudo-labels for each text $x_i \in \mathcal{D}$, subsequently select samples where the pseudo-labels match the original labels $y_i$ to construct the positive datasets $D_p$. Here, we treat the samples in $D_p$ as positive samples, which train the noisy label detector in such a logical form: "input text belongs to this label." The details of LLM prompts are in Appendix \ref{Construction of Prompt}.

\section{Learning From Clean samples and Noisy samples}
\label{Appendix: Loss functions}
In this part, we present the details of the classifier model training to learn from clean samples and noisy samples. Denote $f(x)\in \mathbb{R}^K$ as the output of the classifier model $f$, where $K$ is the number of classes. The confidence of $x$ for each class $k \in \left\{1,\dots,K\right\}$ can be represented as follows: 
$ p(k;x)  =  \frac{e^{f(k;x)}}{\sum_{k=1}^{K} e^{f(k;x)}}$.

\textbf{Learning from Clean Samples.}
For the selected clean samples $x_i \in \mathcal{D}_c$, we directly utilize the cross-entropy loss for the classifier model:

\begin{align}
\label{cross-entropy loss}
\mathcal{L}_{\mathcal{D}_c}=-\frac{1}{N}\sum_{i=1}^{N_{\mathcal{D}_c}} \mathrm{log}p(y_i;x_i)
\end{align}

where $\mathcal{D}_c$ is the selected clean subsets, $N_{\mathcal{D}_c}$ denotes the size of $\mathcal{D}_c$, and $N$ denotes the size of the entire dataset.

\textbf{Learning from Noisy Samples.} After selecting clean samples from $\mathcal{D}$ by the noisy label detector, the remaining samples are considered noisy samples. To maximize the utilization of training data, we 
construct the demonstrations via clean samples and leverage the strong in-context
learning ability of LLMs (GPT-4o)  to generate new labels for these noisy samples. Then, we put these corrected samples in the correction subsets $\mathcal{D}_o$. With the help of LLMs (GPT-4o), the number of noisy samples has greatly decreased. However, even the most powerful LLM cannot generate the right labels for each noisy sample. To learn from the corrected noisy samples in $\mathcal{D}_o$, we resort to the reversed cross-entropy loss function. This loss function has a noise-robust property, which can let us optimize the classifier model given a dataset with a lower noise ratio \cite{DBLP:conf/aaai/GhoshKS17, DBLP:conf/nips/ZhangS18, DBLP:conf/acl/YuanCZJ24}.  Specifically, we utilize the reversed cross-entropy loss for sample ($x_i$, $y_i$) in $\mathcal{D}_o$:

\begin{align}
\label{reversed cross-entropy loss}
 \mathcal{L}_{\mathcal{D}_o}=-\frac{1}{N}\sum_{i=1}^{N_{\mathcal{D}_o}}\sum_{k=1}^{\mathcal{K}}
 p(k;x_i)
 {\rm{\log}} \, q(k|x_i),
\end{align}

where $q(k|\boldsymbol{x})$ is the ground-truth distribution over labels, $N_{\mathcal{D}_o}$ denotes the size of $\mathcal{D}_o$.

\textbf{Overall Learning Objectives.}
Finally, we train the classifier model on selected clean samples and relabeled noisy samples by: $\mathcal{L}=\mathcal{L}_{\mathcal{D}_c}+\mathcal{L}_{\mathcal{D}_o}$.

\section{Detailed Process for Generating Noisy Labels}
\label{Appendix: Detailed Process for Generating Noisy Labels}
High-quality data is typically crucial. However, in real-world scenarios, collected data often contains biases \cite{lv2025debiased} and noise. We have simulated such non-ideal conditions in our experiments. Specifically, in our evaluative experiments, we first select the following datasets: \textbf{Trec}, \textbf{SST-2}, \textbf{SST-5}, \textbf{20ng}, and \textbf{AGNews}. Then we synthesize noisy labels with a variety of noise types and ratios for these datasets. When the noise ratio $ \varepsilon \in[0,1)$ is given, we explain the details of synthetic noise generation processes in the following:

\textbf{Symmetric noise.} 
Symmetric noise chooses one of the other classes at random to replace the label. Each class has the same probability of incorrectly flipping to any other class. To generate this noise, we establish the noise transition matrix $T \in R^{K\times K}$, where $K$ represents the number of classes. We then modify the elements in the noise transition matrix according to the following formula:
\begin{align*}
\begin{split}
T_{ij}= \left \{
\begin{array}{ll}
    \varepsilon,   & i=j\\
    \frac{1-\varepsilon}{k-1},    & i\neq j
\quad ,
\end{array}
\right.
\end{split}
\end{align*}
where $i$ and $j$ respectively represent the horizontal and vertical axes in the noise transition matrix. Lastly, we flip the labels in training samples based on the probability in the matrix.

\textbf{Asymmetric noise.}
Asymmetric noise carries out pairwise label flipping, in which a class $i$ can only change to the following class ($i$ mode $K$) + 1. That is to say, similar classes are mistakenly flipped between each other. To generate this noise, we establish the noise transition matrix $T \in R^{K\times K}$, where $K$ represents the number of classes. We then modify the elements in the noise transition matrix according to the following formula:
\[
T_{ij} =
\begin{cases}
\varepsilon, &  i = j \\
1 - \varepsilon, &  i = j+1 \text{ (mod } K\text{)} \\
0, & \text{otherwise}
\end{cases}
\]
where $i$ and $j$ respectively represent the horizontal and vertical axes in the noise transition matrix. Lastly, we flip the labels in training samples based on the probability in the matrix.

\textbf{Instance-dependent noise.} Instance-dependent noise alters labels according to the transition probability determined by the related attributes of the instance. The generation of such noise is affected by text features, which is more consistent with the noise generation process in the real world, and more challenging. We follow previous works \cite{DBLP:conf/acl/YuanCZJ24} for instance-dependent noise generation. The detailed algorithm of noisy label generation is summarized in Algorithm \ref{Algorithm 1}.

\begin{algorithm*}[ht!]
\caption{ Instance Dependent Noise Generation}
\label{alg:Framwork}
\begin{algorithmic}[1] 
\REQUIRE   
   Clean samples $(x_i,y_i)_{i=1}^n$, $y_i\in[1,\dots,k]$ ;
   Noisy ratio $\varepsilon$;
   
    \STATE Train an LSTM classifier $f$; 
    \STATE Get output from an LSTM classifier $f_{x_i} \in \mathbb{R}^{k}$ for all $i=1,\dots, n$;
    \STATE Set $N_{noisy}=0$;
     \WHILE{$N_{noisy}< n \times \varepsilon$} 
    \STATE Randomly choose a sample $x_i$, $argmax(softmax(f_{x_i}))\neq y_i$;
    \STATE set its noisy label 
    $\bar{y_i} = argmax(softmax(f_{x_i}))$; \\
    \STATE $N_{noisy}=N_{noisy}+1$;
    \ENDWHILE
 \ENSURE  
    Noise samples $(x_i,\bar{y_i})_{i=1}^n$ ;
\end{algorithmic}
\label{Algorithm 1}
\end{algorithm*}

\section{Details of Real-World Datasets with Noisy Labels}
\label{Appendix: Details of Real-World Datasets with Noisy Labels}
In our evaluative experiments, we also select
the following real-world datasets with noisy labels: \textbf{Hausa}, \textbf{Yorùbá}, and \textbf{AlleNoise}. \textbf{Yorùbá} and \textbf{Hausa} are low-resource African languages with five and seven categories, respectively, in their text categorization datasets. With a level of 33.28\% for the latter and 50.37\% for the former, they both incorporate real-world noise. \textbf{AlleNoise} comprises 502310 brief texts categorized by 5692 types. Due to incorrectly categorized data points, there is a 15\% noise level in it.

Table \ref{Table 1} introduces detailed statistics about all datasets used in our experiments.
\begin{table}[th]
\small
\setlength\tabcolsep{4pt}
\centering
\begin{tabular}{l| c c c c}
\toprule[1.3pt]
\makecell[l]{\textbf{\#Dataset}} & \makecell[c]{\textbf{\#Class}} & \makecell[c]{\textbf{\#Training}} & 
\makecell[c]{\textbf{\#Validation}} & 
\makecell[c]{\textbf{\#Test}}\\ 
\midrule
\rm Trec &  6 & 4952 & 500 & 500\\
\rm 20ng &  20 & 9051 & 7527 & 2263\\
\rm AGNews &  4 & 112400 & 7600 & 7600\\
\rm SST-2 &   2 & 5099 & 1820 & 1820\\
\rm  SST-5 &  5 & 8544 &  1101 & 2210\\
\rm  Hausa &  5 &  2045 & 290 & 582\\
\rm  Yorùbá &  7 &  1340 & 189 & 379\\
\rm  AlleNoise &  5692 &  400k & 50k & 50k\\
\bottomrule[1.3pt]
\end{tabular}
\caption{The detailed statistics of all datasets used in our experiments.}
\label{Table 1}
\end{table}

\section{Details of Baselines}
\label{Appendix: Details of Baselines}
In our evaluative experiments, we compare our Delora with the following LNL methods:

\textbf{Basic models.} We train the LLaMA-3.1-8B-Instruct only with standard cross-entropy loss without noise handling.

\textbf{Co-Teaching} \cite{DBLP:conf/nips/HanYYNXHTS18}. Co-Teaching trains two models simultaneously and lets them instruct one another using each mini-batch.

\textbf{SelfMix} \cite{DBLP:conf/coling/QiaoDDLCC022}. SelfMix uses the Gaussian Mixture Model to split samples and semi-supervised learning to manage label noise.

\textbf{SENT}\cite{DBLP:conf/emnlp/WangLWCLZCY22}. SENT transfers the noise distribution to a clean set and trains a model to distinguish noisy labels from clean ones using model-based features.

\textbf{LAFT} \cite{DBLP:conf/emnlp/WangTGL23}. LAFT examines the possibility of using supervision data—such as confidence scores—produced by ChatGPT to address the noisy label issue in pre-trained language model fine-tuning.


\textbf{NoiseAL} \cite{DBLP:conf/acl/YuanCZJ24}. NoiseAL is a novel framework that introduces active learning to combine the non-pretrained model (BiLSTM), pre-trained language model (BERT), and LLMs for learning from noisy labels.

\textbf{CleaR} 
\cite{DBLP:conf/acl/KimK024}. CleaR is a new PEFT technique that minimizes the impact of noisy data while adaptively activating the PEFT modules to enhance generalization ability.

\section{Implementation Details and Setups}
\label{Appendix: Implement Details}
In this section, we detail to implement the baselines and our Delora.

For Delora and all baselines, we report the average performance on 5 different seeds considering their stochasticity. In the main experiments, we chose the 
LLaMA-3.1-8B-Instruct as the backbone model for Delora (both the noisy label detection stage and the model training stage) and other baseline
methods, the LLaMA was fine-tuned by using LoRA.

\paragraph{LoRA implementation.} For Delora, we set the bottleneck deminsion $r$ for the clean LoRA and noisy LoRA as 32. For these two LoRAs, we only apply LoRA weights on query and value attention weights.

\paragraph{Hardware Details.}
We train our framework on Nvidia RTX 3090 and Nvidia A100 GPU. We utilize mixed precision training \cite{DBLP:conf/iclr/MicikeviciusNAD18} to expedite the training procedure. All the implementations are performed with Python, PyTorch, and HuggingFace.
\paragraph{Hyper-parameters.}
 In order to strike a balance between effectiveness and efficiency, we set 8 training epochs for the noisy label detection stage, and 6 training epochs for the classifier model training stage. Moreover, in the noisy label detection stage, we perform model warm-up for 2 epochs on all datasets. We select the batch size from $[16, 32]$, and sweep the learning rates in $[1e-4, 2e-4, 3e-4, 4e-4, 5e-4]$ for Delora. The selection of hyper-parameters is selected according to the performance on a clean development set.

 In our work, our proposed Delora is a two-stage framework consisting of a noisy label detection stage and a classifier model training phase. The pseudo-code is presented in Algorithm \ref{pseudo-code}.

\section{More detailed Results}
\label{Appendix: More detailed Results}
To further demonstrate the broad applicability of our proposed method, we have evaluated the proposed methods on the 20ng and AGNews datasets. Table \ref{Appendix: result of noisy label detection} shows the evaluation results for noisy label detection. Table \ref{Appendix: result of text classification} shows the evaluation results for text classification.

\begin{table*}[htb!]
    \centering
    \small
    \setlength{\tabcolsep}{4.2pt} 
    \begin{tabular}{cc|cccc|cccc|cccc}
        \toprule
        \multirow{2}{*}{\textbf{Dataset}}&
        \multirow{2}{*}{\textbf{Method}} & \multicolumn{2}{c}{20\%S} & \multicolumn{2}{c|}{40\%S} & \multicolumn{2}{c}{20\%A} & \multicolumn{2}{c|}{40\%A} & \multicolumn{2}{c}{20\%I} & \multicolumn{2}{c}{40\%I} \\
        \cmidrule(lr){3-6} \cmidrule(lr){7-10}
        \cmidrule(lr){11-14}
        & & Prec. & Rec. & Prec. & Rec. & Prec. & Rec. & Prec. & Rec. & Prec. & Rec. & Prec. & Rec. \\
        \midrule
        \multirow{3}{*}{\textbf{20ng}} & LLMs-detection &  97.97 & 59.80 & 98.17 & 58.97 & 98.40 & 59.62 & 98.17 & 58.88 & 98.63 & 59.75 & 97.76 & 59.46\\        
        & Small-loss & 81.71 & 86.50 & 63.02 & 83.78 & 81.07 & 88.17& 61.00& 85.97& 81.16 & 86.50 & 61.59 & 80.25\\
     \rowcolor{green!20} \cellcolor{white}{} &\cellcolor{white}{Delora (Ours)}
        &  \textbf{79.68} & \textbf{71.34} & \textbf{60.17} & \textbf{58.43} & \textbf{80.46} & \textbf{72.80} & \textbf{60.04} & \textbf{59.56} & \textbf{96.16} & \textbf{85.14} & \textbf{72.54} & \textbf{86.90}\\   
        \midrule
        \multirow{3}{*}{\textbf{AGNews}} & LLMs-detection & 99.81 & 63.69 & 99.31 & 63.52 & 99.25 & 60.34 & 98.82 & 60.32 &99.45 & 60.34 &98.74 & 53.69 \\
        & Small-loss& 80.82& 92.24&81.09&90.19& 80.95& 91.22&80.61&92.24& 81.09 & 90.19 & 80.95 & 91.22\\
        \rowcolor{green!20} \cellcolor{white}{} &\cellcolor{white}{Delora (Ours)} &\textbf{99.28} & \textbf{93.85} & \textbf{97.70} & \textbf{93.86} & \textbf{99.16} & \textbf{93.94} & \textbf{96.71} & \textbf{94.29} & \textbf{99.36} & \textbf{95.01} & \textbf{97.39} & \textbf{94.85}\\
        \bottomrule[1.0pt]
        
    \end{tabular}
    \vspace{-0.3cm}
    \caption{We evaluate the 
    performance of noisy label detection (\textit{i.e.}, clean sample selection performance) on the 20ng and AGNews datasets by comparing the precision (\%) and recall (\%) of Delora with LLMs-detection and small-loss. The best results results are highlighted in
\textbf{Bold}.}.
    \label{Appendix: result of noisy label detection}
\end{table*}

\begin{table*}[htb!]
     \caption{Performance (test accuracy \%) comparison of Delora with other LNL baselines on synthetic noise datasets (20ng and AGNews datasets). Base (Clean) refers to the base model trained on ground truth data without noisy labels. LLM-base refers to directly using LLMs (GPT-4o)
      on the test dataset. \textbf{Bold} means the best score for each dataset.}
    \label{Appendix: result of text classification}
    \centering
   \renewcommand\arraystretch{1.5}
\setlength\tabcolsep{3pt}
    \begin{tabular}{c |c c c c c c |c c c c c c}
    \toprule
  
\multirow{2}*{\textbf{Model}}  &
    \multicolumn{6}{c|}{\textbf{20ng}}  & \multicolumn{6}{c}{\textbf{AGNews}}  \\  
\cmidrule(lr){2-7}\cmidrule(lr){8-13}
  & 20\%\textbf{S} & 40\%\textbf{S} & 20\%\textbf{A} & 40\%\textbf{A} & 20\%\textbf{I} & 40\%\textbf{I} & 20\%\textbf{S} & 40\%\textbf{S} & 20\%\textbf{A} & 40\%\textbf{A} & 20\%\textbf{I} & 40\%\textbf{I}\\
  \midrule
  Base (Clean)  
& \multicolumn{6}{c|}{88.14}
& \multicolumn{6}{c}{96.16}\\
  Base  & 83.85 & 72.78 & 82.98 & 63.79 & 81.80& 70.12 & 91.46 & 88.39 & 91.71 & 88.88 & 91.62 & 88.57\\
  \midrule
  LLM-base  &  \multicolumn{6}{c|}{72.15} & \multicolumn{6}{c}{84.52} \\
  \midrule
  Co-Teaching & 85.35 & 72.85 & 83.98 & 63.94 & 82.16 & 71.97 & 92.03 & 88.75 & 92.55 & 89.38 & 92.88 & 88.88 \\
\midrule
SENT  & 84.17 & 73.97 & 83.80 & 64.40 & 82.44 & 71.56 & 92.57 & 89.31 & 92.37 & 89.91 & 92.6 & 89.45  \\
\midrule
   LAFT  & 85.64 & 74.17 & 83.95 & 64.37 & 82.64 & 71.58 & 93.09 & 91.38 & 93.19 & 89.82 & 92.81 & 90.65\\
  \midrule  
   SelfMix  & 80.87 & 78.99 & 78.19 & 65.52 & 77.68& 70.54 & 92.22 & 89.45 & 92.88 & 90.65 & 91.97 & 89.24 \\
  \midrule
   CleaR  & 84.58 & 72.89 & 83.28 & 64.87 & 83.58 & 70.44 & 92.98 & 90.30 & 92.76 & 89.39 & 92.46 & 89.80\\
  \midrule
   NoiseAL  & 85.95 & 77.11 & 85.89 & 75.79 & 84.62 & 75.69 &  92.20 & 90.31 & 93.08 & 89.02 & 92.69 & 90.31\\
  \midrule
  \rowcolor{green!20}  \cellcolor{white}{Delora (Ours)} &  \textbf{88.51}& \textbf{82.16}& \textbf{88.40}& \textbf{79.60}& \textbf{86.87}& \textbf{80.90}& \textbf{95.64} &\textbf{95.29}&\textbf{95.84}&\textbf{95.17}&\textbf{95.64}&\textbf{95.86}\\
  \midrule[1.0pt]
    \end{tabular}
      \vspace{-0.2cm}
\end{table*}

\section{Construction of Prompt}
\label{Construction of Prompt}
In this section, we list the prompt used in our experiments.
To optimize the noisy label detector, we leverage the LLM in the construction of positive datasets. The prompt is as follows:
\begin{tcolorbox}[colback=black!5!white, colframe=black, title=Zero-Shot-CoT Prompt]

Below is a text classification problem. 
Note that you can only select the label in \textcolor{blue}{\{options\}}.  
Let's think step by step and give your answer.

SENTENCE: \textcolor{blue}{\{text\}}  

LABEL:   
\end{tcolorbox}

Then, in the training of classifier models, we use the selected clean samples to construct demonstrations, prompting LLM to generate the new labels via Few-Shot-CoT. The prompt is as follows:

\begin{tcolorbox}[colback=black!5!white, colframe=black, title=Few-Shot-CoT Prompt]

Below is a text classification problem. 
Note that you can only select the label in \textcolor{blue}{\{options\}}.  
Let's think step by step and give your answer.

SENTENCE: \textcolor{blue}{\{text1\}}  

LABEL: \textcolor{blue}{\{label1\}}  

SENTENCE: \textcolor{blue}{\{text2\}}  

LABEL: \textcolor{blue}{\{label2\}} 

\ldots  

SENTENCE: \textcolor{blue}{\{text\}}  

LABEL:   
\end{tcolorbox}

\section{More Detailed Analysis}
\label{More Analysis}

\subsection{Results under Extreme Noise Conditions.} 
To further validate the robustness of the proposed methods under extreme noise conditions, we evaluate our model on Trec datasets under large noise ratios. Table \ref{Table: the result of noisy label detection on Trec under 60} reports the precision and recall for noise detection, where Delora exhibits a robust capability in identifying severe noise. 
Table \ref{Table: the result of classification on Trec under 60} reports the test accuracy for text classification, where Delora exhibits a robust capability in classifying texts. Compared to other baselines, our methods show superior performance under high noise ratios.

\begin{table}[htb!]
    \centering
    \small
    \setlength{\tabcolsep}{4.0pt} 
    \begin{tabular}{cc|cccc}
        \toprule
        \multirow{2}{*}{\textbf{Dataset}}&
        \multirow{2}{*}{\textbf{Method}} & \multicolumn{2}{c}{60\%S} & \multicolumn{2}{c}{60\%A} \\
        \cmidrule(lr){3-6} 
        & & Prec. & Rec. & Prec. & Rec. \\
        \midrule
        \multirow{3}{*}{\textbf{Trec}} & LLMs-detection & 60.26 & 60.32 & 60.04 & 60.05\\        
        & Small-loss & 59.14 & 73.45 & 58.58 &  69.76\\
        \rowcolor{green!20} \cellcolor{white}{} &\cellcolor{white}{Delora (Ours)} &  \textbf{97.75} & \textbf{92.97} & \textbf{97.04} & \textbf{92.25} \\   
        \bottomrule[1.0pt]
        
    \end{tabular}
    \vspace{-0.2cm}
    \caption{We evaluate the 
    performance of noisy label detection (\textit{i.e.}, the clean sample selection performance) on the Trec by comparing the Precision (\%) and Recall (\%) of Delora with LLMs-detection and Small-loss.}
    \label{Table: the result of noisy label detection on Trec under 60}
\end{table}

\begin{table*}[thb!]
\footnotesize
\setlength\tabcolsep{7pt}
\centering
\begin{tabular}{l r| c c c c c c c c c c}
\toprule[1.0pt]
\makecell[l]{\textbf{Dataset}} & &  \multicolumn{8}{c}{\textbf{Trec}} \\ 
\midrule
\textbf{Noise}($\downarrow$) / \textbf{Method}($\rightarrow$)  &     & 
   Base  & 
   Co-Teaching & 
SENT &  LAFT &  SelfMix & CleaR & NoiseAL & Delora \\
\midrule
60\%S & & 78.40 & 80.06 & 81.28 & 81.46 & 81.52 & 79.93 & 81.12 & \textbf{84.53}\\
\midrule
60\%A & & 56.20 & 61.27 & 62.25 & 63.68 & 65.36 & 65.52 & 67.21 & \textbf{71.96}\\
\bottomrule[1.0pt]
\end{tabular}
\caption{The detailed results (test accuracy \%) on Trec datasets. \textbf{Bold} means the best score.
}
\label{Table: the result of classification on Trec under 60}
\end{table*}

\begin{table*}[h!]
\setlength\tabcolsep{7pt}
\centering
\scalebox{1.0}
{
\begin{tabular}{c c c|c c c c}
\toprule[1.0pt]
\multicolumn{3}{c}{\textbf{Modules}}& \multicolumn{4}{c}{\textbf{Trec}} \\
\midrule
\makecell[c]{\textbf{dual LoRAs}} & \makecell[c]{\textbf{constraint on} $\Delta \sigma_n$}
& \makecell[c]{\textbf{constraint on} $\Delta \sigma_c$}
& \makecell[c]{\textbf{20\%S}} & \makecell[c]{\textbf{40\%S}}& \makecell[c]{\textbf{20\%A}}& \makecell[c]{\textbf{40\%A}}\\ 
\midrule

\XSolidBrush & \XSolidBrush
& \XSolidBrush & 95.20 & 90.20 & 94.20 &  87.40  \\
 \midrule
 
 \CheckmarkBold & \XSolidBrush 
& \XSolidBrush & 82.27 & 77.82 & 80.96 & 75.63 \\
 \midrule
 
 \CheckmarkBold & \CheckmarkBold 
& \XSolidBrush & 96.37 & 95.21 & 95.86 & 95.01\\
 \midrule

\CheckmarkBold & \CheckmarkBold 
& \CheckmarkBold &  \textbf{98.46} & \textbf{97.60} & \textbf{98.30}& \textbf{97.40} \\

\bottomrule[1.0pt]
\end{tabular}
}
\caption{Further ablation study for the noisy label detector on the Trec dataset.}
\label{Table:Further Ablation}
\end{table*}

\begin{figure*}[h!]
  \centering
    \subfigure[The performance of our methods with different $h_2$ under different noise scenarios]{\includegraphics[width=0.48\textwidth,height=0.4\textwidth]{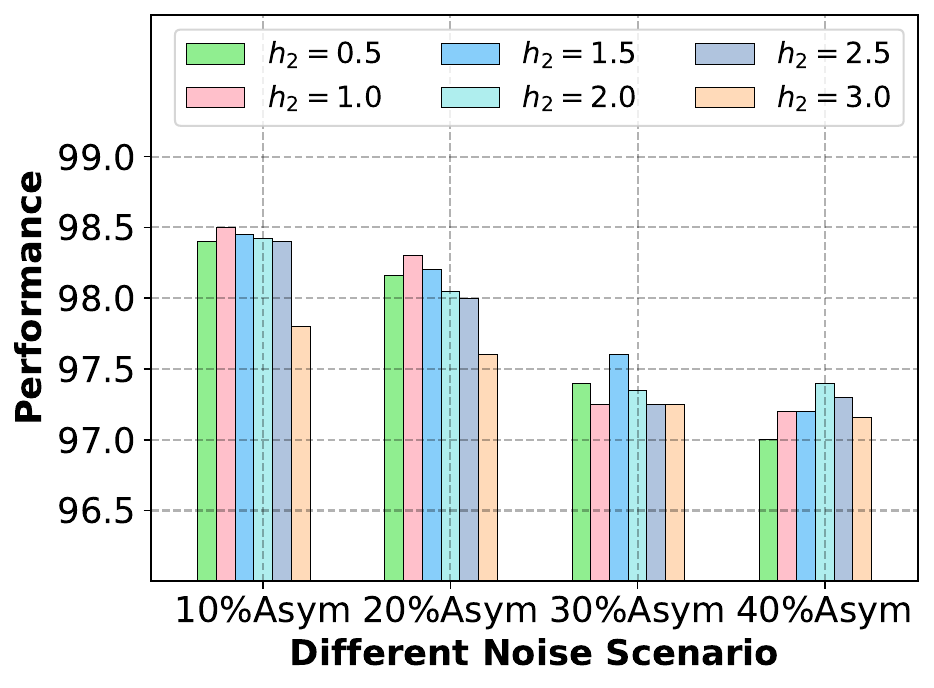}} 
    \subfigure[The performance of our methods with different $h_2$ under different noise scenarios]{\includegraphics[width=0.48\textwidth,height=0.4\textwidth]{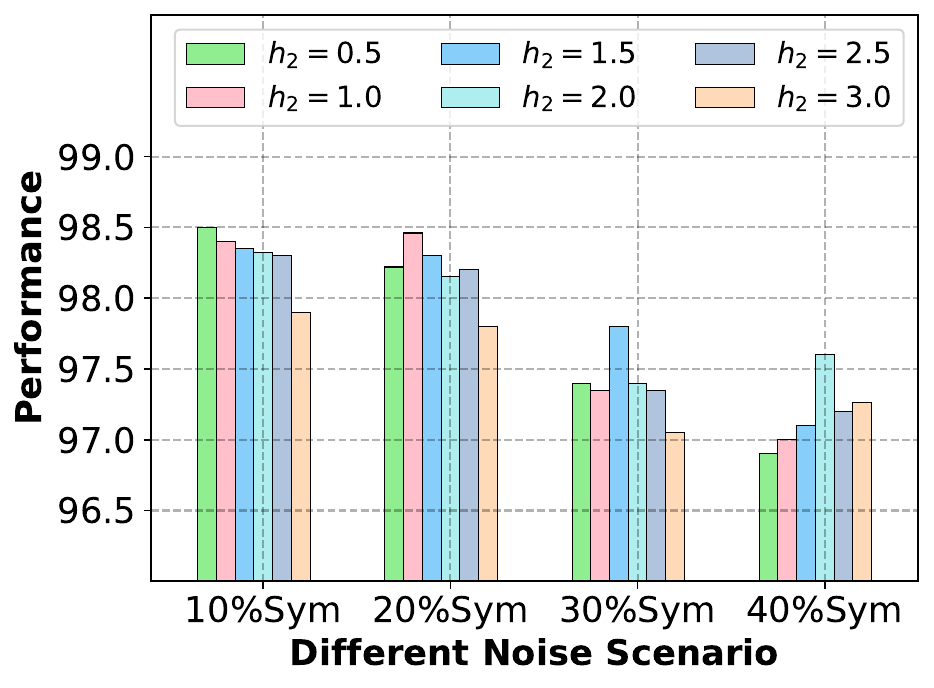}} 
    \subfigure[Relationship between the optimal $h_2$ and noise ratio]
{\includegraphics[width=0.48\textwidth,height=0.4\textwidth]{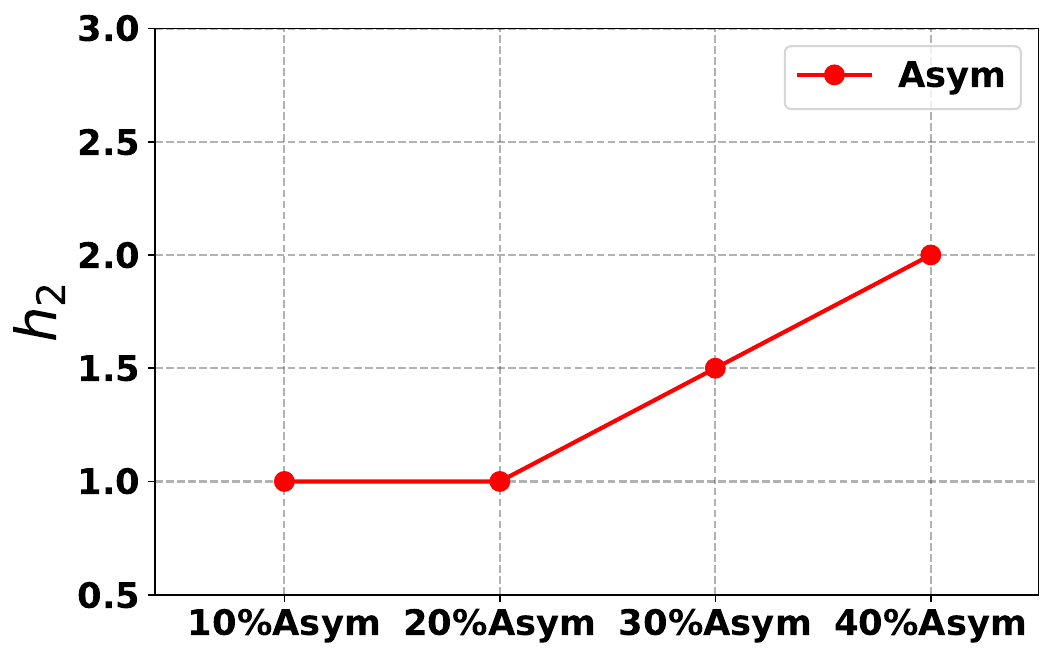}} 
    \subfigure[Relationship between the optimal $h_2$ and noise ratio]{\includegraphics[width=0.48\textwidth,height=0.4\textwidth]{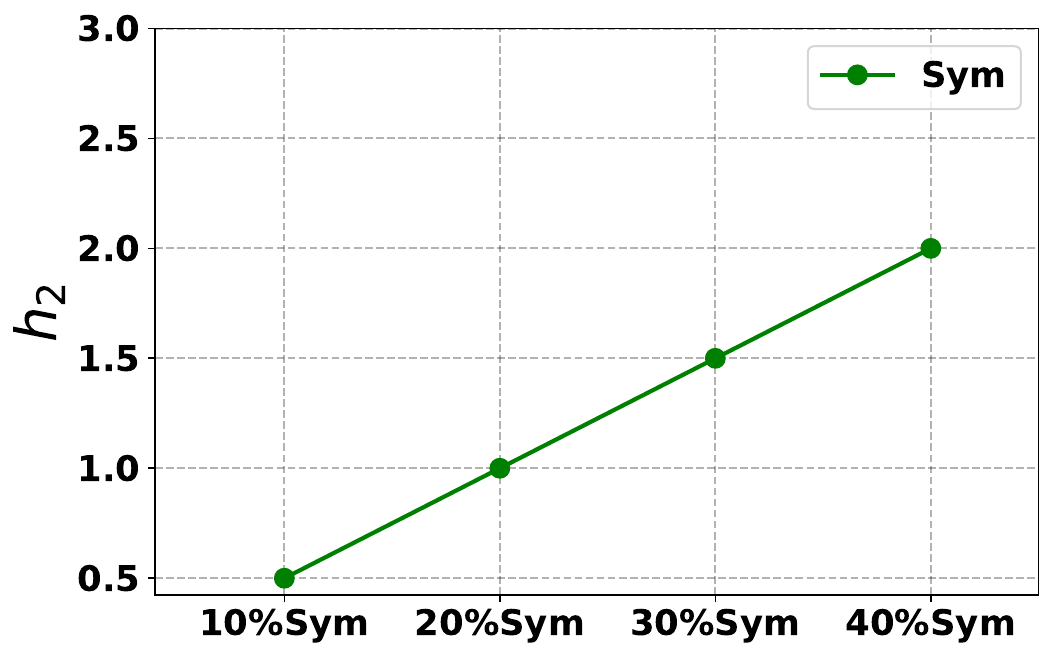}} 
  \caption{Analysis for the choice of hyper-parameter $h_2$ under different noise ratios on the Trec datasets.}
    \label{Analysis for the choice of hyper-parameter h2 under different noise ration}
\end{figure*}

\subsection{Further Analysis for the modules in Noisy Label Detector.}
\label{Further Analysis for the modules in Noisy Label Detector.}
As outlined in Section \ref{Identifying Noisy Labels with Dual LoRAs}, our proposed noisy label detector consists of three key modules: the dual LoRA (
the clean and noisy LoRA), a constraint on the parameter update \( \Delta \sigma_c \) of the clean LoRA, and a constraint on the parameter update \( \Delta \sigma_n \) of the noisy LoRA. In this part, we conduct further ablation studies to elucidate the factors that contribute to the success of our approach. Experiments are performed on Trec datasets, and experimental results are shown in Table \ref{Table:Further Ablation}. 

Firstly, we only introduce two LoRA modules and do not constrain their parameter update, it can be seen that the classification performance on the test set is very poor. Since dual LoRA modules are simultaneously influenced by noisy labels during training, they fail to learn distinct representations for clean and noisy samples. As a result, they struggle to effectively identify noisy samples in the noisy label detection phase, ultimately hindering the training of the downstream classifier model. That is to say, simply introducing more LoRA modules itself will not bring any performance improvement and even have negative impacts. 

Secondly, if we add the constraint on the parameter update $\Delta \sigma_n$ of noisy LoRA, the test accuracy is significantly improved. This suggests that imposing a constraint on $\Delta \sigma_n $ allows the noisy LoRA to absorb the adverse effects of mislabeled data, enabling the clean LoRA to focus on learning from clean samples more effectively. That is to say, it can help different LoRAs to learn distinct representations for clean and noisy samples.

Thirdly, we introduce an additional constraint on the parameter update \( \Delta \sigma_c \) of the clean LoRA. The results show that this further enhances test accuracy, indicating that restricting \( \Delta \sigma_c \) helps the clean LoRA retain less mislabeled data, leading to improved performance.

Overall, further studies show that combining the three modules of our proposed noisy label detector can achieve a steady performance improvement, which proves the necessity of each module in the noisy label detector.

\subsection{Further Analysis for the Hyper-parameter Setting in Constraint Functions.}
As we mentioned in Section \ref {Further Analysis for the modules in Noisy Label Detector.}, we add a constraint on the parameter update \( \Delta \sigma_c \) of the clean LoRA $\Delta w_c$, and a constraint on the parameter update \( \Delta \sigma_n \) of the noisy LoRA $\Delta w_n$. Specifically, in this paper, we set $\tau_1(t) = t^{h_1}$ and $\tau_2(t) = t^{-h_2}$ to constrain the parameter update of clean LoRA and noisy LoRA, where $h_1$ and $h_2$ are two hyperparameters. Here, we explore the hyper-parameter setting in these two constraint functions.

\paragraph{Hyper-parameter $h_2$.}
 Specifically, in Figure \ref{Analysis for the choice of hyper-parameter h2 under different noise ration} (a), we compare the performance of different \( h_2 \) values under various Asym noise conditions. For each noise setting, we identify the optimal \( h_2 \) value and visualize the results in Figure \ref{Analysis for the choice of hyper-parameter h2 under different noise ration} (c). Similarly, in Figure \ref{Analysis for the choice of hyper-parameter h2 under different noise ration} (b), we compare the performance of different \( h_2 \) values under various Sym noise conditions. For each noise setting, we identify the optimal \( h_2 \) value and visualize the results in Figure \ref{Analysis for the choice of hyper-parameter h2 under different noise ration} (d).  
 
 From the results in Figure \ref{Analysis for the choice of hyper-parameter h2 under different noise ration} (c-d), we observe a positive correlation between the optimal \( h_2 \) value and the noise ratio. In other words, as the noise ratio increases, a larger \( h_2 \) accelerates the updates of \( \Delta w_n \), allowing the noisy LoRA \( \Delta w_n \) to better absorb the impact of mislabeled data. 
 
\begin{figure*}[h!]
  \centering
    \subfigure[The performance of our methods with different $h_1$ under different noise scenarios]{\includegraphics[width=0.48\textwidth,height=0.4\textwidth]{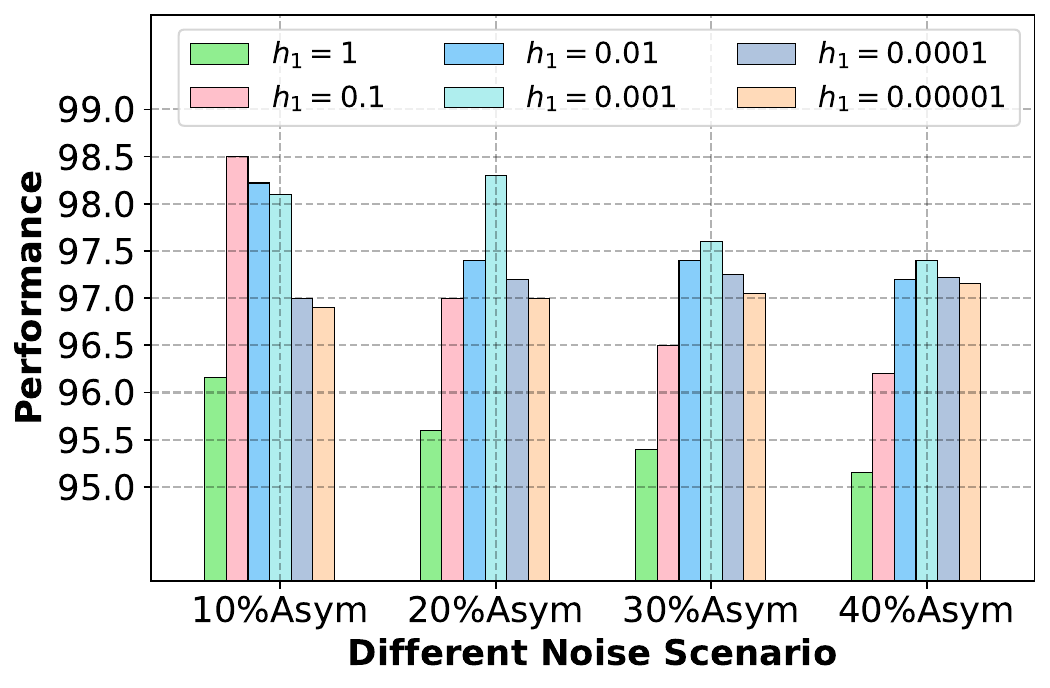}} 
    \subfigure[The performance of our methods with different $h_1$ under different noise scenarios]{\includegraphics[width=0.48\textwidth,height=0.4\textwidth]{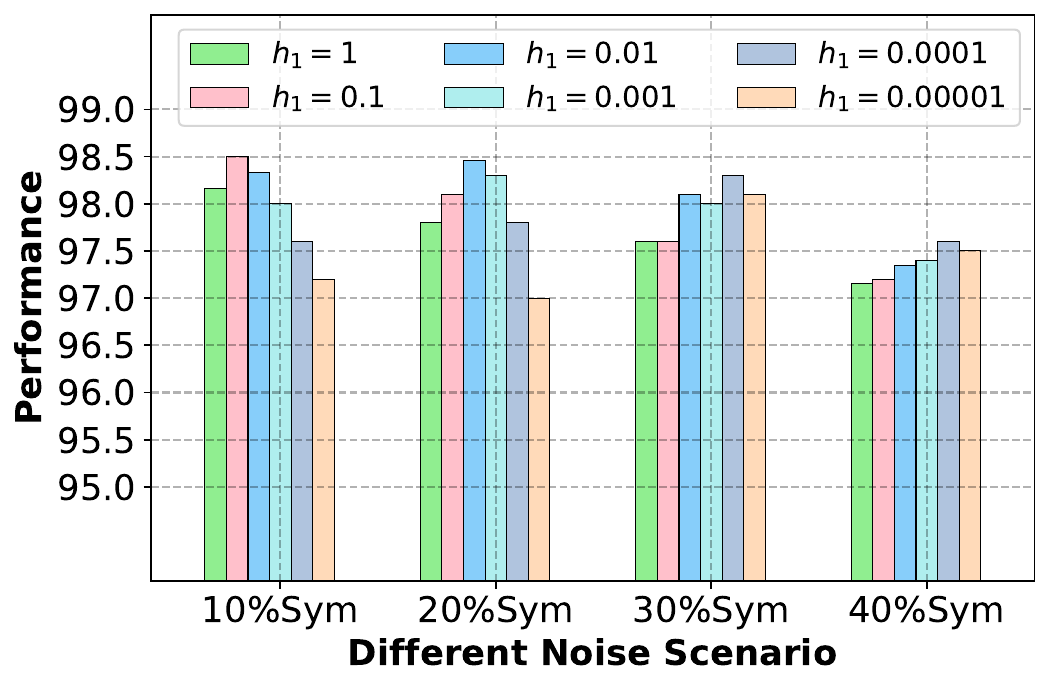}} 
    \subfigure[Relationship between the optimal $h_1$ and noise ratio]
{\includegraphics[width=0.48\textwidth,height=0.4\textwidth]{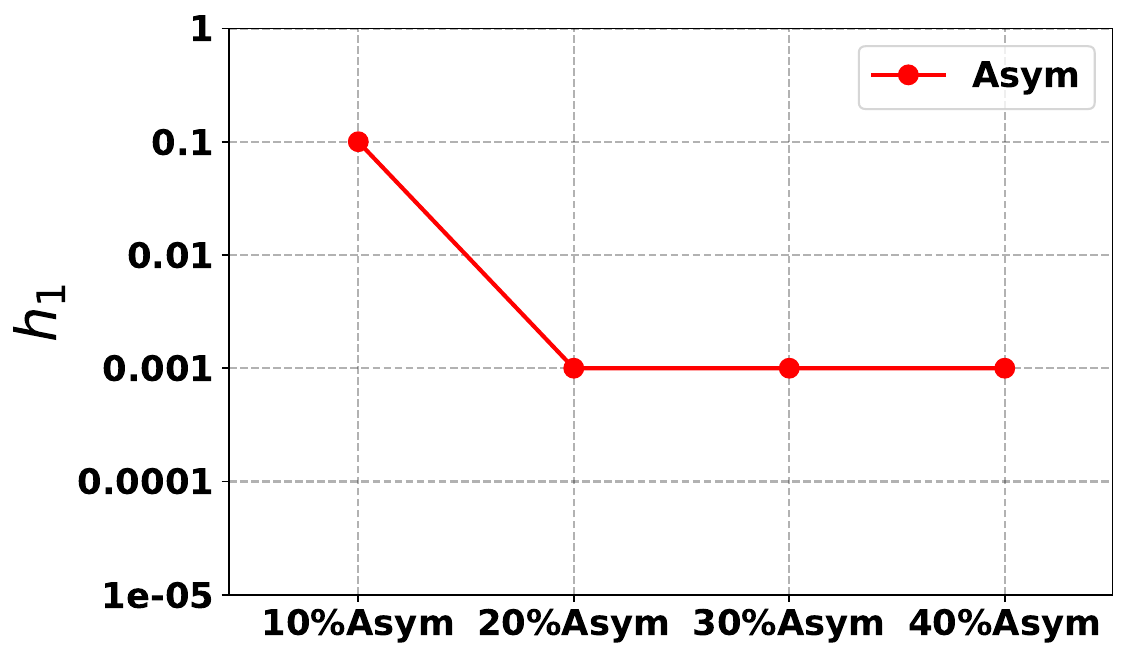}} 
    \subfigure[Relationship between the optimal $h_1$ and noise ratio]{\includegraphics[width=0.48\textwidth,height=0.4\textwidth]{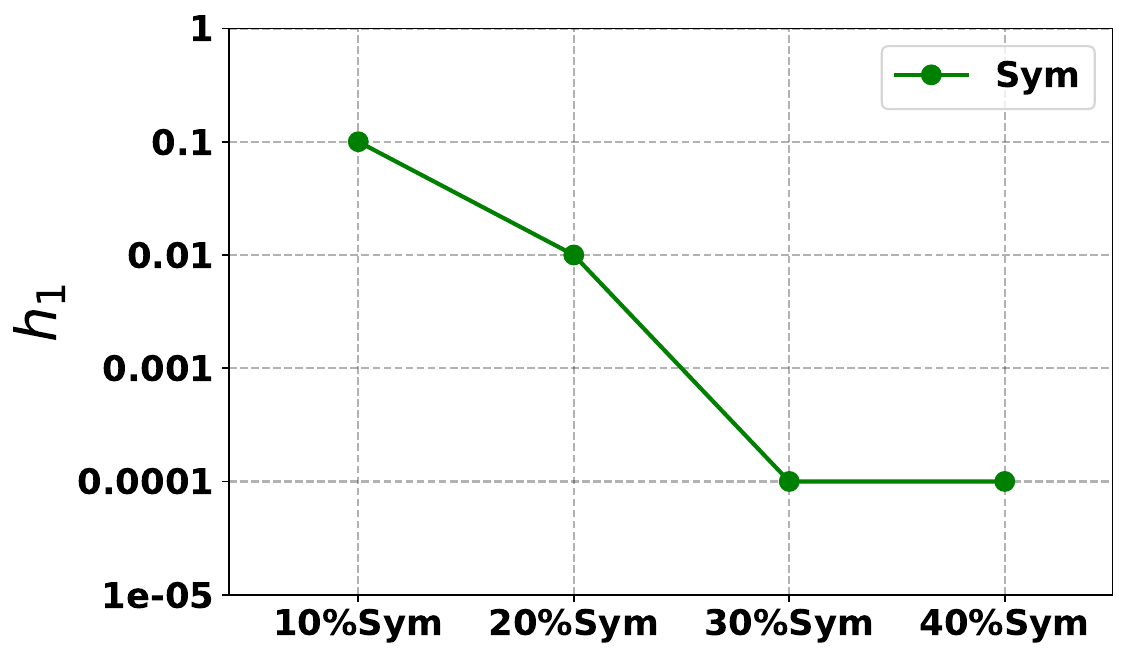}} 
  \caption{Analysis for the choice of hyper-parameter $h_1$ under different noise ratios on the Trec datasets.}
    \label{Analysis for the choice of hyper-parameter h1 under different noise ration}
\end{figure*}

 \paragraph{Hyper-parameter $h_1$.}
 In Figure \ref{Analysis for the choice of hyper-parameter h1 under different noise ration} (a), we compare the performance of different \( h_1 \) values under various Asym noise conditions. For each noise setting, we identify the optimal \( h_1 \) value and visualize the results in Figure \ref{Analysis for the choice of hyper-parameter h1 under different noise ration} (c). Similarly, in Figure \ref{Analysis for the choice of hyper-parameter h1 under different noise ration} (b), we compare the performance of different \( h_1 \) values under various Sym noise conditions. For each noise setting, we identify the optimal \( h_1 \) value and visualize the results in Figure \ref{Analysis for the choice of hyper-parameter h1 under different noise ration} (d).  

From the results in Figure \ref{Analysis for the choice of hyper-parameter h1 under different noise ration} (c-d), we observe a negative correlation between the optimal \( h_1 \) value and the noise ratio. That is to say, as the noise ratio increases, a smaller \( h_1 \) accelerates the updates of \( \Delta w_c \), allowing the clean LoRA \( \Delta w_c \) to better prevent the impact of mislabeled data. 

 Overall, adding the constraint on the parameter update \( \Delta \sigma_c \) of the clean LoRA and on the parameter update \( \Delta \sigma_n \) of the noisy LoRA 
can make different LoRAs focus on learning different information from clean and noisy samples, thereby increasing the robustness of the noisy label detector.


\begin{algorithm*}[ht!]
\caption{The proposed framework Delora}
\begin{algorithmic}[1] 
\REQUIRE   
   A training dataset $\mathcal{D}$=$\left\{(x_i,y_i)\right\}^{N}_{i=1}$, $y \in
   \left\{1,\dots,K\right\}$,
   warmup epochs $T_w$,
   epochs $T_d$ to train the noisy label detector, epochs $T_c$ to train the classifier model, clean LoRA parameters $\Delta w_c$, noisy LoRA parameters $\Delta w_n$.
    \STATE \textcolor{gray}{// Stage 1: Training the Noisy Label Detector}

    \WHILE{epoch $\leq$ $T_w$}
    \STATE Warm-up the LLM containing dual LoRAs on the $\mathcal{D}$ by $\mathcal{L}_{warm} = \mathcal{L}_{ce} + \mathcal{L}_{LoRA}$
    \STATE epoch = epoch + 1 
    \ENDWHILE

    \WHILE{epoch $\leq$ $T_d$}
    \STATE Construct the clean subset $D_c$ by Eq. (\ref{eq:threshold}) and Eq. (\ref{eq-phi})
    \STATE Construct the positive and negative samples
    \STATE Update parameters of $\Delta w_c$ and $\Delta w_n$ by  $\mathcal{L} = \mathcal{L}_{ce}+\mathcal{L}_{LoRA} +\mathcal{L}_{Detector}$
    \STATE epoch = epoch + 1; 
    \ENDWHILE    
    \STATE \textcolor{gray}{// Step 2: Training the Classifier Model}
    \WHILE{epoch $\leq$ $T_c$}
    \STATE Compute the cross-entropy loss $\mathcal{L}_{\mathcal{D}_c}$for selected clean samples by Eq.(\ref{cross-entropy loss})
    \STATE Compute the reversed cross-entropy loss $\mathcal{L}_{\mathcal{D}_o}$for relabeled noisy samples by Eq.(\ref{reversed cross-entropy loss})
    \STATE Update the parameter of the classifier model by $\mathcal{L}=\mathcal{L}_{\mathcal{D}_c}+\mathcal{L}_{\mathcal{D}_o}$
    \STATE epoch = epoch + 1 
    \ENDWHILE    
\end{algorithmic}
\label{pseudo-code}
\end{algorithm*}

\subsection{Effect of Robust Loss Function}
\label{Effect of Robust Loss Function}
As discussed in Section \ref{Appendix: Loss functions}, due to GPT-4o being unable to generate correct labels for each noisy sample selected by the noisy label detector, we utilize the reversed cross-entropy loss functions to better learn from $\mathcal{D}_o$ with a certain noise ratio. We conduct an ablation experiment (see Table \ref{An ablation experement for robust loss function}) to verify the effectiveness of this robust loss function by replacing it with cross-entropy loss functions.

\begin{table*}[thb!]
\setlength\tabcolsep{2.5pt}
\centering
\scriptsize
\begin{tabular}{l r| c c c c| c c c c c c| c c c c c c}
\midrule[1.0pt]
\makecell[l]{\textbf{Dataset}} & & \multicolumn{4}{c|}{\textbf{Trec}} & \multicolumn{6}{c|}{\textbf{SST-5}}& \multicolumn{6}{c}{\textbf{SST-2}} \\ 
\midrule
\textbf{Loss function}($\downarrow$) / \textbf{Noise}($\rightarrow$)  &     & 
20\%\textbf{S} & 40\%\textbf{S} & 20\%\textbf{A} & 40\%\textbf{A} & 20\%\textbf{S} & 40\%\textbf{S} & 20\%\textbf{A} & 40\%\textbf{A} & 20\%\textbf{I} & 40\%\textbf{I} & 20\%\textbf{S} & 40\%\textbf{S} & 20\%\textbf{A} & 40\%\textbf{A} & 20\%\textbf{I} & 40\%\textbf{I}\\
\midrule
cross-entropy loss
& & 97.21 & 96.1 & 96.98 & 95.98 & 56.34 & 54.04 & 56.32 & 53.41 & 55.39 & 53.59 & 94.6 & 94.07 & 94.51 & 93.33 & 94.73 & 93.74\\
\midrule
reversed cross-entropy loss
 & & \textbf{98.46} & \textbf{97.60} & \textbf{98.30}& \textbf{97.40}& \textbf{57.39}& \textbf{55.62}& \textbf{57.57}& \textbf{55.39}& \textbf{57.02}& \textbf{55.02}& \textbf{96.50} &\textbf{95.75}&\textbf{96.27}&\textbf{95.18}&\textbf{96.08}&\textbf{95.00}\\
\midrule[1.0pt]
\end{tabular}
\caption{
Ablation study for loss functions on  $\mathcal{D}_o$.  \textbf{Bold} means the best score.
}
\label{An ablation experement for robust loss function}
\end{table*}

\end{document}